\documentclass{article}

\usepackage{amsmath,amsfonts}

\usepackage{algorithmic}
\usepackage{algorithm}

\usepackage{array}
\usepackage[caption=false,font=normalsize,labelfont=sf,textfont=sf]{subfig}
\usepackage{textcomp}
\usepackage{stfloats}
\usepackage{url}
\usepackage{verbatim}
\usepackage{graphicx}
\usepackage{cite}
\usepackage{amssymb}
\usepackage{booktabs}
\usepackage{xcolor}
\newtheorem{assumption}{Assumption}[section]
\newtheorem{theorem}{Theorem}[section]

\newtheorem{lemma}[theorem]{Lemma}

\begin{document}

\title{Neural Network-Based Change Point Detection for Large-Scale Time-Evolving Data}

\author{Jialiang Geng,
    George Michailidis
    \thanks{First Author is with the Department of Statistics, University of California, Los Angeles (email: cauchygeng@g.ucla.edu).}% 
    \thanks{Second Author is with the Department of Statistics, University of California, Los Angeles (email: gmichail@g.ucla.edu).}%
}

\markboth{Journal of \LaTeX\ Class Files,~Vol.~14, No.~8, August~2021}%
{Shell \MakeLowercase{\textit{et al.}}: A Sample Article Using IEEEtran.cls for IEEE Journals}


\maketitle

\begin{abstract}
The paper studies the problem of detecting and locating change points in multivariate time-evolving data. The problem has a long history in statistics and signal processing and various algorithms have been developed primarily for simple parametric models. In this work, we focus on modeling the data through feed-forward neural networks and develop a detection strategy based on the following two-step procedure. In the first step, the neural network is trained over a prespecified window of the data, and its test error function is calibrated over another prespecified window. Then, the test error function is used over a moving window to identify the change point. Once a change point is detected, the procedure involving these two steps is repeated until all change points are identified. The proposed strategy yields consistent estimates for both the number and the locations of the change points under temporal dependence of the data-generating process. The effectiveness of the proposed strategy is illustrated on synthetic data sets that provide insights on how to select in practice tuning parameters of the algorithm and in real data sets. Finally, we note that although the detection strategy is general and can work with different neural network architectures, the theoretical guarantees provided are specific to feed-forward neural architectures.
\end{abstract}


\section{Introduction}
The problem of change point detection has a rich history across various disciplines, including statistics, econometrics, signal processing, and machine learning, owing to its wide range of applications in areas such as quality control \cite{Lai}, economics, finance, and risk analysis \cite{ZHU201518}, medical diagnosis \cite{LIU201849}, social network analysis \cite{Kendrick2018ChangePD}, and linguistics \cite{Kulkarni:langchange,Wang2018RealtimeCP}. 
Change point detection methods are typically categorized according to the task in hand: \textit{online} methods aim to detect changes in the data distribution/dynamics, as soon as they occur based on streaming data, while \textit{offline} methods aim to \textit{retrospectively} detect changes, when all data are available for analysis. The former task is also referred to in the literature as event or anomaly detection, while the latter task is also sometimes called signal segmentation.  

Various detection algorithms have been developed for a number of \textit{parametric} models in the offline setting, including exhaustive search \cite{Braun2000MultipleCF}, penalized cost approaches \cite{Harchaoui2010MultipleCE,Killick_2012} and approaches based on screening and ranking criteria \cite{Niu_2012}. Many of these detection algorithms also come with probabilistic guarantees on the number and locations of the change points detected (Asymptotic consistency) for specific parametric models, such as mean shift time series models \cite{fryzlewicz2014wild}, vector autoregressive models \cite{PeiliangBai,SafikhaniBM22}, regression type models \cite{bai1998estimating,article}, random graph and stochastic block network models \cite{sto,HaoChen}. Meanwhile, the evaluation of the different detection algorithms is also important. The paper \cite{Truong_2020} offers a comprehensive overview of algorithms for offline change point detection in multivariate time series and discusses consistency for these methods including Asymptotic consistency and evaluation metrics including Hausdorff distance and F1 score, many of which will also be used in this paper.

The rapid advancement and widespread use of neural network architectures have highlighted the need for developing change point detection methods tailored specifically to these models, as most existing approaches are designed for parametric models. Current literature on this topic is relatively sparse and primarily focuses on online change point detection. For instance, \cite{titsias2020sequential} proposed a sequential detection procedure with a predefined detection window, utilizing a generalized likelihood ratio test where the likelihood is approximated by a simple feed-forward network. Similarly, \cite{probratio} reformulated the detection problem as a classification task, employing neural network architectures to train a classifier. In \cite{Ebrahimzadeh2018PyramidRN}, change point detection is performed using a novel deep neural network (DNN) structure called the Pyramid Recurrent Neural Network, which combines recurrent neural networks (RNNs) and convolutional neural networks (CNNs). The output is then passed through a binary classifier to determine the presence of a change point. Another approach, described in \cite{Bulunga2012ChangepointDI}, integrates singular spectral analysis with auto-associative neural networks, using the residuals to detect change points based on a specified threshold.
Lastly, \cite{GUPTA2022118260} combines preprocessing steps such as normalization and recursive singular spectral analysis with a neural network-based autoencoder. The reconstruction error from the autoencoder serves as the criterion for identifying change points under a predefined threshold.
%where the null hypothesis will be rejected if drastic changes are found in the probability ratio of the model under normal assumptions for the score function. Another paper\cite{probratio}, instead of detecting the changes in probability ratio directly, developed a labeling method that transformed the original problem into one classification problem with neural networks. These methods either require additional assumptions like normality\cite{titsias2020sequential} or are hard to be extended to other types of neural networks like CNN or RNN\cite{probratio}. Moreover, these methods often apply a prespecified detection window but without instructions to select the window size and lack the consistency analysis of the locations and number of the change points detected.\\

 On the other hand, to the best of our knowledge, change point detection methods for \textit{nonparametric offline setting} time evolving data mainly focused distribution change instead of time evolving mechanism change. Paper \cite{kernelmodel} proposed a kernel-based model selection framework that minimizes a penalized kernel least-squares criterion enabling the detection of an unknown number of change points. The optimal offline time segmentation method in \cite{optimalseg} innovatively combines dynamic programming, cost-based optimization, and error guarantees to offer a robust solution. Random Forests for Change Point Detection in \cite{randomforestschangepoint} proposed a nonparametric method leveraging machine learning classifiers, such as random forests, to construct a classifier log-likelihood ratio which is used to detect multiple change points in time series data. However, the deep learning based methodologies are sparse. The paper \cite{spectrum} proposed a mSSA-based CPD algorithm for detecting change points in multivariate time series and focused on the spectral change instead of the traditional mean change or variance change. Existing deep learning methods including KLCPD \cite{klcpd} used kernel-based hypothesis testing and deep generative models to mitigate insufficient samples in the abnormal distribution and the paper \cite{Li_2024} combined the CUSUM Statistics and deep learning. These paper mainly focused on the change of distribution instead of change of time evolving mechanism and many of the literature does not include the theoretical analysis of Asymptotic consistency. Meanwhile, some literature including \cite{randomforestschangepoint} and \cite{spectrum} only tackle the single change point problem. Hence, the goal of this work is to develop a detection algorithm for feed-forward networks focusing in time evolving data that comes with probabilistic guarantees on identifying the correct number and locations of multiple change points. It is based on a two-step procedure and leverages a nonparametric criterion function.

Based on the few available references above, emphasize differences in the methodology, in addition to theory in the next paragraph.
The main contribution of this paper are 
\begin{enumerate}
    \renewcommand{\labelenumi}{\Roman{enumi}.} % Roman numeral for level 1
    \item Developed a nonparametric two-step framework leveraging feed-forward neural networks and a systematic approach to work on both single change point case and multiple change points case in nonparametric multidimensional time-evolving data which is illustrated by algorithm \ref{alg:error} and \ref{alg:detect}
    \item The algorithm works in both the regression problem demonstrated in section \ref{sec:problem} equation \eqref{eq:problem} and time evolving data like Vector Autoregressive (VAR) model.
    \item Theorems in section\ref{sec:thm} provided rigorous theoretical analysis and guarantees for accuracy in detecting the number of change points and bounding the distance between true and estimated change points. The theoretical analysis varies in different cases of the dataset including temporal independent cases in theorem \ref{thm:independent} and dependent dataset in theorem \ref{thm:dependent} and subgaussian dataset in theorem \ref{thm:subgaussian}.
    \item The effectiveness of the proposed
    strategy is illustrated on synthetic data sets that provide insights
    on how to select in practice tuning parameters of the algorithm
    and in real datasets. 

\end{enumerate}

The remainder of the paper is organized as follows: Section\ref{sec:problem} sets up the problem under consideration and introduces the nonparametric criterion function. Section \ref{sec:detection} describes the two-step algorithm, while Section \ref{sec:thm} provides the theoretical guarantees for it. Finally, Section \ref{sec:exp} presents numerical experiments based on simulated data that showcase the role that various assumptions play in the performance of the algorithm, while Section \ref{sec:exp} illustrates the algorithm on selected real-world data sets.

\section{Problem Formulation}
\label{sec:problem}
The setting under consideration is as follows:
Let \(Y_i, i=1,\cdots,T_{sum}\) be a response/output vector of dimension \(h\) and \(X_i, i=1,\cdots,T_{sum}\) a vector of dimension \(p\) of covariates/attributes, related through the following predictive model:
\begin{equation}
\label{eq:problem}
Y_i=f_j(X_i)+\epsilon_i,\ \ \tau_{j-1}<i\leq \tau_j 
\end{equation}
where \(\tau_0\leq i\leq \tau_{N+1}, \tau_0=0, \tau_{N+1}=T_{sum}\) correspond to the N \textit{change points} and \(T_{sum}\) is the number of total time points available. It can be seen that the model \(f_j\) is applicable for a segment of time points and changes to another one \(f_{j+1}\) at change point \(\tau_j+1\). 
%For this regression model, the \(X_i\) is the multivariate input data with dimension d at time point i, and \(Y_i\) is the output which is also multivariate with dimension h. 
The model function \(f_j, \mathbb{R}^p\rightarrow \mathbb{R}^h\) is assumed to be measurable and composed of several functions (technical details provided in the Supplement). This framework can also cover the Vector Autoregressive (VAR) model:
\begin{equation}
Y_i=A_{1j}Y_{i-1}+...+A_{qj}Y_{i-q}+\epsilon_i,\tau_{j-1} \leq i\leq \tau_j
\end{equation}
where \(Y_i\in \mathbb{R}^h\) and \(A_{ij}\in \mathbb{R}^{h\times h},1\leq i\leq q \) is the transition (regression coefficient) matrix. If we treat the q lags of \(Y_i\) as \(X_i=Y_{i-q\leq k<i}\), then \(Y_i=\bar{f}_j(X_i)+\epsilon_i\) where \(\bar{f}_j=\bar{f}_j(A_{ij})\) which can be covered by the proposed framework.(Technical details can be found in section \ref{sec:exp}).

\textbf{Criterion function:}
The following criterion function is employed in the detection algorithm:
\(E(t)=\sum_{i=t}^{T_2+t}||Y_i-\hat{f}(X_i)||^2\), where a fully connected feed-forward network \(\hat{f}:\mathbb{R}^{p_0}\to \mathbb{R}^{p_{L+1}}\) is defined as: \(\hat{f}=W_L\sigma_{v_L}W_{L-1}\sigma_{v_{L-1}}...W_1\sigma_{v_1}W_0x\) and \(L\) is the number of layers of the network, \(W_j\) is the \(p_{j+1} \times p_j\) matrix, \(v_j \in \mathbb{R}^{p_j}\) and \(\sigma_{v_j}(x)=\max(0,x-v_j)\) is the ReLU function with bias \(v_j\). Here \(p_0=p\) and \(p_{L+1}=h\). \(||.||\) is the 2-norm of a vector or matrix. The criterion function measures the squared error loss between the actual and fitted values and mimics analogous criteria used in change point analysis for traditional parametric statistical models. Further,  the neural network is trained over a certain time period (to be specified in section \ref{sec:detection}), and then the criterion function is evaluated over a window of size \(T_2\), whose starting point is a generic point in time t. The length of the time window \(T_2\) is also specified in the section \ref{sec:detection}.

In this paper, we used this neural network architecture represent \(\hat{f}\) and estimate the true underlying non-linear predictive model \(f_j\) which can also be estimated by other regression model including kernel regression in other literature. 

\section{Detection Algorithm}
\label{sec:detection}
\textbf{Annotations and assumptions:} The \(T_{sum}\) is total number of time points available as defined in section \ref{sec:problem}. \(T_1,T_2,T_3\) refer to the training window size, test window size and detection window size. \(\pi\) refers to the detection threshold.

\begin{assumption}
\label{asp:signal}
Define the signal of the model change \(M1_i\) and \(M1_i\leq \mathbb{E}[||f_{i+1}(X)-f_{i}(X)||^2]\leq M2_i<\infty\) where \(f_i\) is model function defined in equation \ref{eq:problem} and \(M2_i\) is some constant and define \(M1_*=\min_iM1_i\) as the minimum signal of the model change. We assume
\begin{equation}
M1_*>4h\sigma^2
\end{equation}
\end{assumption}
This assumption ensure the signal of the model change is large enough compared to noise so the algorithm will be able to detect the change point. 

\begin{assumption}
\label{asp:distance}
Define \(T=\min_{i=0,...,N}|\tau_{i+1}-\tau_i|\) as the minimum distance between change points. We assume it has a lower bound:
\begin{equation}
T\geq \frac{C_0 b(T_{sum})}{M1_*}
\end{equation}
for some constant \(C_0\) and function b. Usually, we use \(b(T_{sum})=\sqrt{T_{sum}}\)
\end{assumption}
This assumption provide a lower bound for the distance between true change points and a function of the relationship between this distance and model change signal. Such balance between signal change and distance between true change points are very common in literature like the papers in review \cite{Truong_2020}.

Recall that in the assumption, \(\tau_0=0,\tau_{N+1}=T_{sum}\). This assumption also restrict the change points should be distant enough from the boundaries as well which can let the model pretrained on the data before the first change point and retrained after the change point. This requirement is also very common and mild in change point detection and the classical method always require the change point \(cT_{sum}\leq\tau\leq (1-c)T_{sum}\) for some constant c. Then the search for change point starts after a time interval proportional to \(T_{sum}\).

\begin{assumption}
\label{asp:smoothness}
Each model function \(f_j=g_q\circ g_{q-1}...\circ g_1 \circ g_0 \) where each \(g_i\) is a \(\beta\)-H\"older smoothness function.
\end{assumption}
This assumption guaranteed the smoothness of the model function which give the neural network capability of approximate the true model accurately. The detailed definition of \(\beta\)-H\"older smoothness function will be in the supplement section \ref{subsec:betaholder}.

The proposed change point detection algorithm is a two-step procedure, wherein the first step the Error Criterion function \(E(t)\) is computed for each time point t from 1 to \(T_{sum}\)  and in the second step, its maximum point is identified. We will present this algorithm in next two subsections \ref{subsec:single} and \ref{subsec:single} for both single change point case and multiple change points case.

\subsection{Single change point}
\label{subsec:single}
\begin{figure}[!t]
\centering
\includegraphics[width=3.2in]{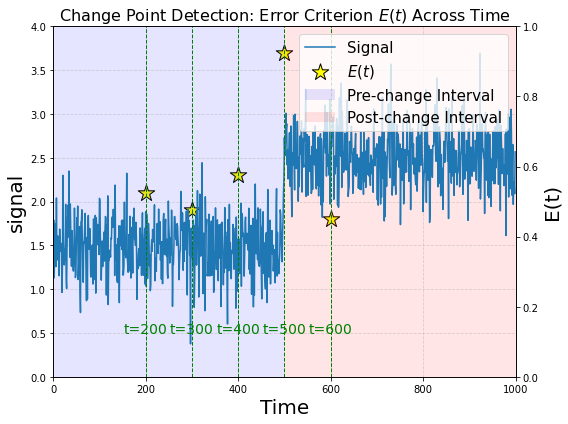}
\caption{Example for generating E(t)}
\label{fig:toy_error}
\end{figure}

It would always be easier to detect the change point if there is only one. To provide intuition, consider the following simple univariate ``signal plus noise'' model, 
\begin{equation}
    y_i=\alpha I(i<\tau)+\beta I(i\geq \tau)+\epsilon_i
\end{equation}
with a single change point \(\tau\). Before the change point, the output \(y_i\) takes the value \(\alpha=1\) corrupted by random noise \(\epsilon_i\), while after the change point it becomes \(\beta=2\) also corrupted by random noise \(\epsilon_i\). The total observation interval is of length \(T_{sum}=1000\) and the change point is located in the middle of the interval at \(\tau=500\). The generated data with the signal levels \(\alpha\) and \(\beta\) are shown in the left panel of Figure \ref{fig:toy_error}. Suppose that a simple feed-forward neural network is trained on the interval (100,200) with training window size \(T_1=100\). The error function is then evaluated over a testing window of length \(T_2=100\) from 201 to 300 at selected time points over the observation interval, which in this case is 200, 400, and so forth until a change point is detected. The spacing between the evaluation points is a tuning parameter of the algorithm. It can be seen that at $t=200$, the value of the error function E(t)\ is low as depicted by the first star in the figure \ref{fig:toy_error}. The results are similar for the windows starting at $t=300$ and $400$  indicated by the second and third stars in the plot. Since the signal changes at $t=500$, while \(\hat{f}\) is trained on the first model, while the testing window spans on the interval 500 to 600, the Error criterion will exhibit a large value, since the responses \(Y_i\) are generated by the second model, while the fitted values are based on the first model. Later at time point 600, the neural network is trained on data from the 500-600 observations which are from the second model and the Error criterion is evaluated also on the second model and will drop to a low level depicted by the fifth star.

The first step of the algorithm generating the error function E(t) is presented in algorithm \ref{alg:error}. Given a time point t, let \(X[t-T_1:t],Y[t-T_1:t]\) be the input and output from time \(t-T_1\) to \(t\) and with the dimension \(T_1\times p, T_1\times h\) and a feed forward network \(\hat{f}_{\theta}\) is fitted in this input and output and \(E(t)\) is calculated by the test error in time period between \(t\) and \(t+T_2\).
\begin{algorithm}[H]
\caption{Test Error Generator}
\label{alg:error}
\begin{algorithmic}[1]
    \STATE \textbf{System Initialization}
    \STATE Input data \(X_t\), prediction \(Y_t = f_i(X_t) + \epsilon_t, \; i = 1, 2, \ldots\), training window size \(T_1\), test window size \(T_2\), maximum training step \(T\), neural network function \(\hat{f}_{\theta}\) with trainable parameter \(\theta\).
    \FOR{\(t = T_1, 2, \ldots, T\)}
        \STATE Compute the loss: \[ \mathcal{L}_{\theta} = \| Y[t-T_1:t] - \hat{f}_{\theta}(X[t-T_1:t]) \|^2 \]
        \WHILE{\(\theta\) not converged}
            \STATE Update parameters: \( \theta \leftarrow \text{ADAM}(\mathcal{L}_{\theta}) \)
        \ENDWHILE
        \STATE Record test error:
        \[
        E[t] = \| Y[t:t+T_2] - \hat{f}_{\theta}(X[t:t+T_2]) \|^2
        \]
    \ENDFOR
    \STATE \textbf{Output} test error function \(E(t)\)
\end{algorithmic}
\end{algorithm}
The detection step for single change point case is simple, we can select the time point t that maximize the function \(E(t)\) as the estimated change point which is \(t=500\) in this example. The consistency of the estimation can be guaranteed by theorem \ref{thm:single}.
%\begin{figure*}[!t]
%    \centering
%    \subfloat[ Example for generating E(t)]{\includegraphics[width=0.5\textwidth]{toy-error_mod.png}} 
%    \subfloat[Example for detection process]{\includegraphics[width=0.5\textwidth]{detect_mod.png}} 
%    \caption{A simple example}
%\label{fig:exp}
%\end{figure*}

\begin{figure}[!t]
\centering
\includegraphics[width=3.2in]{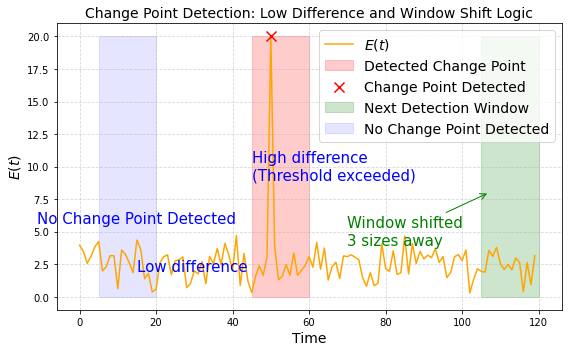}
\caption{Example for detection process}
\label{fig:detect}
\end{figure}

\subsection{Multiple Change points}
\label{subsec:multiple}
The detection starts to get complicated when there are unknown number of multiple change points in the given data. Then the selection of window size become important because the algorithm will miss some change points if they are in the same training or testing window. And the detection algorithm need to be modified as well. We introduce a detection window with the tuning parameter detection window size \(T_3\) which also need to be carefully selected for the same reason. We also introduce detection threshold \(\pi\) as we need to differentiate the cases when detection window contains change point and when it does not. The detailed selection strategies of these window size and detection threshold will be discussed in the section \ref{subsec:selection} and \ref{sec:thm}.

In modified detection step, for each time point t, we select a detection window \((t-T_3,t+T_3)\) span at the center of this point with a designated detection window size and calculate the difference between the maximum and minimum values of \(E(t)\) within this window. If the difference exceeds a fixed threshold \(\pi\) (appropriately selected), we detect one change point and the location will be the time point maximizing the \(E(t)\) within this detection window and the detection window can be moved to the next location without overlapping with previous detection window. The technical results established in the next section \ref{sec:thm}, provide guarantees of how well estimated the true change point is and also provide guidance of how to select the various window sizes required in steps 1 and 2 and of the detection algorithm.

For example, consider an existing test error with one peak at time point 50 in the figure \ref{fig:detect}, the peak threshold \(\pi=4\), and fix the detection window size \(T_3=15\). At the time point before 44, within the detection window, the difference between the maximum and minimum of the error function is less than \(\pi\) as shown by the first blue arrow, so we move to the next time point. At the time point between 44 and 54, which is near the peak, the difference will, however, exceed the threshold as displayed by the second blue arrow, and then we will set the time point maximizing the test error within the window to be the detected change point. Once upon detecting one change point, we will move on to the time point where the left boundary for the detection window spanned is one window size after the right boundary for the detection window. This strategy will prevent detecting the same change point multiple times which is discussed in the next section.

The detailed strategy is described in algorithm \ref{alg:detect}
\begin{algorithm}[H]
\caption{Multiple Change Point Detection}
\label{alg:detect}
\begin{algorithmic}[1]
    \STATE \textbf{System Initialization}
    \STATE Input error function \( E(t) \), detection window size \( T_3 \), \( n = 1 \), detection threshold \( \pi \), total time points \( T_{\text{sum}} \).
    \FOR{\( t = 1, 2, \ldots, T_{\text{sum}} \)}
        \STATE Calculate 
        \[
        c_t = \max_{\max(t - T_3, 0) \leq i, j \leq \min(t + T_3, T_{\text{sum}})} |E(i) - E(j)|
        \]
        \IF{\( c_t \geq \pi \)}
            \STATE \( \hat{\tau}_n = \arg\max_{\max(t - T_3, 0) \leq i, j \leq \min(t + T_3, T_{\text{sum}})} E(i) \)
            \STATE \( n = n + 1 \)
            \STATE \( t = t + 3T_3 \)
        \ENDIF
    \ENDFOR
    \STATE \textbf{Output:} \( \hat{\tau}_1, \ldots, \hat{\tau}_N \).
\end{algorithmic}
\end{algorithm}
\subsection{Selection of the window size and detection threshold}
\label{subsec:selection}
Given model defined in equation \ref{eq:problem} with total time points \(T_{sum}\), the single change point case is simple as we can directly select \(t=arg\max_tE(t)\). However, when it comes to multiple change points case, it is essential to select the appropriate window size for training, testing, and detection purposes. To simplify the selection process, we can set the training, testing and detection window size \(T_1=T_2=T_0,T_3=2T_0\) for some window size \(T_0\).  Similar to the assumption that ensure distance between change point and the boundary, we also require the distribution of the change points within the total time points \(T_{sum}\) not to be too dense. In other words, there should be a lower bound of the distance between change points that ensure there will not be multiple change points in the same window and the neural network can have enough time points to get optimized. Then the window size \(T_0\) controls upperbound of the number of change points we can detect and the number \(N\leq \frac{T_{sum}}{T_0}\). The details will be discussed further in theorems \ref{thm:independent},\ref{thm:subgaussian} and \ref{thm:dependent}. Given the datasets with appropriate distribution of change points location, for independent and identically distributed data with finite variance, a good choice for window size is \(T_0=\sqrt{T_{sum}}\). For datasets with higher order moments in output space or subgaussian distributed, a narrower window size \(T_0\) can be used. On the other hand, for time-dependent data sets, a larger window size is required (see details in Section \ref{sec:thm}). The number of change points can be detected will also change according to the required window size. In general, when the selection of window size can be smaller, the distribution of change points can be denser and vice versa. In multiple change point case,  We also need a large enough detection threshold \(\pi\) to differentiate the detection window with and without change point. The threshold should be less than the minimum of model change signal \(M1_*\), we use \(\pi=\frac{1}{3}M1_* \) here. And this indicate another assumption that the \(M1_*\) need to be large enough to be differentiated from the random error when there is no change point in the detection window. The detailed configuration will be discussed in \ref{thm:independent} in Section \ref{sec:thm}.

\section{Theoretical Analysis}
\label{sec:thm}
%In this section, we will conduct a consistency analysis for the change points detected. First of all, we will check the validity of the two-step algorithm we discussed in the previous section.
To obtain guarantees for the estimated change point locations, results in \cite{Schmidt_Hieber_2020,ma2022theoretical} for independent and dependent data are leveraged regarding how well the estimated feed-forward network approximates the true underlying \(f_j\).

%estimated feed-forward network approximates the true underlying \(f_j\) for both independent and dependent dataleverage the following papers provides the lemma with a rate of how well the \cite{Schmidt_Hieber_2020},\cite{ma2022theoretical}. With the lemma in these papers, we can prove the theorems below. The proof detail and usage of the lemma will be posted in the supplement section.
Further, a key assumption required is that difference in successive models \(\mathbb{E}[||f_{j}(X)-f_{j-1}(X)||^2]\) should be large enough; otherwise, there will not be adequate signal to detect the corresponding change point.

The next result focuses on establishing properties for the proposed detection algorithm.

\begin{theorem}
\label{thm:single}
%Consider \(Y_i=f_0(X_i)+\epsilon_i, i<\tau\) and \(Y_i=f_1(X_i)+\epsilon_i,i\geq \tau\), where \(Y_i\in \mathbb{R}^h, X_i\in \mathbb{R}^p, var(\epsilon)=\sigma^2\) 
Consider Single change point cases in equation \ref{eq:problem}
and assumption \ref{asp:signal} and \ref{asp:smoothness} hold. %\(M_1=\mathbb{E}[||f_1(X)-f_0(X)||^2]\gg h\sigma^2>0\) with \(M_1\) finite and \(\tau\) denoting the true change point. 
Further, consider the following Error criterion: \(E(t)=\sum_{i=t}^{t+T_0}||\hat{f}(X_i)-Y_i||^2\), with \(T_0\) being the sample size and test size with the neural network \(\hat{f} \)  trained on \(t-T_0\leq i\leq t\).\\
Then, there exists a \(t_0\), such that \(E(t_0)-E(t)>\pi\) defined in section \ref{sec:detection} for all t.\\
Further, let \(\hat{\tau}=argmax_{t}E(t)\) and \(\tau\) denoting the true change point. Then, there exists a constant \(C\), such that
\begin{equation}
|\hat{\tau}-\tau|\leq \frac{CT_0}{M_1-2h\sigma^2}
 \end{equation}
 with probability 
\begin{equation}
1-\frac{C_l}{T_0^{l-1}}
\end{equation}
for some constant \(C_l\) based on l if \(\mathbb{E}[||f_i(X)||^l]<\infty, l\geq 2\)\\
and probability
\begin{equation}
1-C_1e^{-C_2T_0}
\end{equation} 
for some constant \(C_1\) and \(C_2\), if the \(\hat{f},f_j\) is subgaussian for data X.

\end{theorem}

The theorem \ref{thm:single} establishes that step 1 of the detection algorithm 
will identify a detectable peak near the true change point \(\tau\). In addition, the estimated change point \(\hat{\tau}\) detected in step 2 of the algorithm will be "close" to the true change point. Hence, And so once detecting the true change point, the next detection window which will be one detection window size away from the previous one will not contain the previous true change point. As a result, one change point will only be detected for one time.

Next, we consider the case of multiple change points and provide probabilistic guarantees for the proposed detection algorithm regarding the estimated number and locations of them.
%considering the primary purpose of this paper, for offline change point detection, this section will provide the consistency analysis and theoretical approximation of the location and number of the detected change point. The primary purpose of the consistency analysis in this paper is to provide the guarantee that the probability of detecting all the change points accurately in both numbers and locations will be close to 1 if certain assumptions are satisfied.
\begin{theorem}
\label{thm:independent}
Consider independent and identically distributed input data \(X\) and consider the model in equation \eqref{eq:problem}, and assumptions \ref{asp:signal}, \ref{asp:distance} and \ref{asp:smoothness} hold 
%let \(M1_i\leq \mathbb{E}[||f_{i+1}(X)-f_{i}(X)||^2]\leq M2_i<\infty \) and \(M1_*=\min_{i=0,...,N}M1_i\) to be minimum signal and \(T=\min_{i=0,...,N}|\tau_{i+1}-\tau_i|\) minimum distance between change points. 
Let \(\mathbb{E}[||f_i(X)||^l]<\infty\) for $l>2$. Choose a constant \(\rho \in (0,1)\) and define function in assumption \ref{asp:distance} \(b(T_{sum})=\sqrt[\rho(l-1)]{T_{sum}}\)
%Further, assume that \(M1_*>4h\sigma^2\) and for some \(\rho \in (0,1) \), if the assumption holds:
%\begin{equation}
% M1_*T\geq C_0 	\sqrt[\rho(l-1)]{T_{sum}}
%\end{equation}

Then, the algorithm \ref{alg:error} and \ref{alg:detect} with \(T_0 =  T^{\rho} = (\frac{C_0}{M1_*})^{\rho}\sqrt[l-1]{T_{sum}} \) for large \(l\), detection window size \(2T_0\) and detection threshold described in section \ref{sec:detection}, as \(T_{sum}\to \infty\).
satisfies
\begin{equation}
P\bigg(\hat{N}=N,\max_{i=1,...,N}M1_*|\tau_i-\hat{\tau}_i| \\
<C\frac{M1_*}{M1_*-2h\sigma^2}T_{sum}^{\frac{1}{l-1}}\bigg) \to 1
\end{equation}
\end{theorem}

This theorem establishes that, with high probability, the number of detected change points will match the number of true change points. Additionally, distances between the estimated change points and the true change points can be bounded with high probability as well. Notably, the theorem assumes a lower bound on the spacing between consecutive change points. This assumption is critical for determining the maximum number of detectable change points within the given framework, given the total number of observed time points \(T_{sum}\).

For sub-Gaussian datasets, this theorem can be extended to provide a narrower error bound for the distances between the estimated and true change points. Additionally, it enables the detection of a denser and more change points given the total number of time. This extension leverages the properties of sub-Gaussian input data to refine the theoretical guarantees and enhances the performance of the detection algorithm.

\begin{theorem}
\label{thm:subgaussian}
Consider independent and identically distributed input data \(X\) and the model defined in equation \eqref{eq:problem} with subgaussian output and assumptions \ref{asp:signal}, \ref{asp:distance} and \ref{asp:smoothness} hold 
%with \(M1_i\leq \mathbb{E}[||f_{i+1}(X)-f_{i}(X)||^2]\leq M2_i<\infty \) and \(M1_*=\min_{i=0,...,N}M1_i\) and \(T=\min_{i=0,...,N}|\tau_{i+1}-\tau_i|\). Assume that \(M1_*>4h\sigma^2\) and for some \(\rho \in (0,1) \), if
Choose a constant \(\rho \in (0,1)\) and define function in assumption \ref{asp:distance} \(b(T_{sum})=\sqrt[\rho]{log(T_{sum})}\)
%\begin{equation}
% M1_*T\geq C_0 	\sqrt[\rho]{log(T_{sum})}
%\end{equation}

%Further, for \(f_i(X)\) being subgaussian for each i, 
then with the algorithm \ref{alg:error} and \ref{alg:detect} and select \(T_0= T^{\rho} =(\frac{C_0}{M1_*})^{\rho}\log(T_{sum})\), detection window size \(2T_0\) and detection threshold described in section \ref{sec:detection}, as \(T_{sum}\to \infty\).

\begin{equation}
P\bigg(\hat{N}=N,\max_{i=1,...,N}M1_*|\tau_i-\hat{\tau}_i|< \\ 
C\frac{M1_*}{M1_*-2h\sigma^2}log(T_{sum})\bigg)\to 1
\end{equation}

\end{theorem}

In many applications, the datasets may not be independently distributed and may exhibit temporal dependencies. Then, the extension of the theorem to this case is given below.

\begin{theorem}
\label{thm:dependent}
Consider dependent and identically distributed input data \(X\) and the model defined in equation \eqref{eq:problem} and assumptions \ref{asp:signal}, \ref{asp:distance} and \ref{asp:smoothness} hold 
% let \(M1_i\leq \mathbb{E}[||f_{i+1}(X)-f_{i}(X)||^2]\leq M2_i<\infty \) and \(M1_*=\min_{i=0,...,N}M1_i\) and \(T=\min_{i=0,...,N}|\tau_{i+1}-\tau_i|\). 
Let \(\mathbb{E}[||f_i(X)||^2]<\infty\) and \(|cov(f_i,f_j)|<c<1,i\neq j\) for some constant c. 

Choose two constant \(\rho,g \in (0,1)\) and define function in assumption \ref{asp:distance} \(b(T_{sum})=\sqrt[g\rho+1]{T_{sum}}\)
%Assume that \(M1_*>4h\sigma^2\) and for some \(\rho \in (0,1) \) and \(g\in (0,1)\)
%\begin{equation}
% M1_*T\geq C_0 	\sqrt[g\rho+1]{T_{sum}}
%\end{equation}

Then, with the algorithm \ref{alg:error} and \ref{alg:detect} and select \(T_0=T^{\rho}= (\frac{C_0}{M1_*})^{\rho}(T_{sum})^{\frac{\rho}{g\rho+1}}\), detection window size \(2T_0\) and detection threshold described in section \ref{sec:detection}, as \(T_{sum}\to \infty\).

\begin{equation}
P\bigg(\hat{N}=N,\max_{i=1,...,N}M1_*|\tau_i-\hat{\tau}_i|< \\
C\frac{M1_*}{M1_*-2h\sigma^2}T_{sum}^{\frac{\rho}{1+g\rho}}\bigg)\to 1
\end{equation}

\end{theorem}
\subsection{Remarks}
In the assumptions of theorems \ref{thm:independent},\ref{thm:subgaussian} and \ref{thm:dependent}, different characteristics of the data sets requires different lower bounds of the minimum distances between true change points or the density of the change points.  For the independent datasets with the finite \(l-\)momentum of the model output with sufficiently large l, the lower bound of minimum distances for consecutive change points can scale as \(T_{sum}^{\frac{1}{\rho(l-1)}}\). Consequently, the number of detectable change points, given a fixed \(T_{sum}\) scales as \(T_{sum}^{1-\frac{1}{\rho(l-1)}} \). For sub-Gaussian data sets, the lower bound can approach the scale of  \(\log T_{sum}\) enabling the detection of a greater number of change points for a fixed \(T_{sum}\). In contrast, for dependent input data, the lower bound is not less than \(T_{sum}^{\frac{1}{2}}\) and limiting the number of detectable change points to at most \(\sqrt{T_{sum}}\).

\section{Experiments}
\label{sec:exp}
In this section, we present several experiments to illustrate the performance of the proposed algorithm and examine the sensitivity of the results based on the posited assumptions underlying the theoretical results.

\subsection{Synthetic Data}

We focus on experiments with synthetic data and examine the following three settings: the first two involve responses and predictors $(Y_i, X_i)$ generated by different nonlinear mechanisms, while the third one time series data, where the predictors $X_i$ correspond to past lags of the response variable.

%Before analyzing real-world datasets, we will present some experiments on simulation datasets to evaluate the performance of our algorithm and the assumptions underlying the theorems. The first two datasets—feed-forward network data and random forest data are based on regression scenarios in original model equation\eqref{eq:problem}, differing only in the model function \(f_j\).

Following the regression model in Equation  \eqref{eq:problem}, the input data \(X_i\) is generated from i.i.d normal distribution N(0,I) and the dimension of input \(X_i\) is 400 and the dimension of output \(Y_i\) is 200. The error term \(\epsilon_i\) is generated from i.i.d normal distribution N(0,\(I\sigma^2\)) where I is the identity matrix and \(\sigma=0.4\).

For the feed-forward network data, Each \(f_j\) model used a neural network with three layers. In each layer, the fully connected weight matrix is given by \(W_i+W_{ij}\) with no bias vector and activated by the ReLU function. Here \(W_i\) is a fixed matrix for each model function in layer $i$ and \(W_{ij}\) is a sparse matrix generated uniquely for each model function \(f_j\) in layer \(i=1,2,3\). The dimension of the first layer matrix \(W_1\) and \(W_{1j}\) is \(400\times 100\), the dimension of the second layer matrix \(W_2\) and \(W_{2j}\) is \(100\times 100\), and the dimension of the third layer matrix is \(100\times 200\). Each element in \(W_i\) is generated from normal distribution N(0,1), while in \(W_{ij}\) is generated by a sparse distribution with nonzero element following the normal distribution: Sparse(0.1)*N(0,1) where Sparse(0.1)=1 with probability 0.1 and 0 with probability 0.9.

Each \(f_j\) in the Random forest data is generated by a different random forest model with input dimension 400 and output dimension 200. The generating process of each random forest model \(f_j\) is: Given input \(X_i\) and a nonlinear function \(g:\mathbb{R}^{400}\rightarrow \mathbb{R}^{200}\)(Here in our experiments, we used the feed-forward network with weight matrix \(W_i+W_{ij}\) defined in the previous paragraph as g). Let \(\tilde{Y_i}=g(X_i)\) and random forest model \(f_j=arg\min_{f}||f(X_i)-g(X_i)||^2\). And similarly, the output \(Y_i=f_j(X_i)\).

The third data set as previously mentioned, is generated from Vector Autoregression (VAR) model with $q$ lags, as follows:
%is slightly different, the Vector Autoregressive with $p$ lag or VAR(p) data model which is given by equation:
\begin{equation}
\label{eq:VAR}
Y_t=A_{1j}Y_{t-1}+...+A_{qj}Y_{t-q}+\epsilon_t,\tau_{j-1} \leq t\leq \tau_j
\end{equation}
where each \(Y_t\) is a 200 dimensional vector and \(A_{ij}\in \mathbb{R}^{200\times 200},1\leq i\leq q \) is the transition (regression coefficient) matrix. This model can be put into the form of the nonlinear regression model in \eqref{eq:problem} as follows: flatten the predictors (input) as \(X_t=(Y_t,Y_{t-1},...,Y_{t-q+1}), \boldsymbol{\epsilon}_t=(\epsilon_t,...,\epsilon_{t-q+1})\), we have \(Z_t=B_jZ_{t-1}+\boldsymbol{\epsilon}_t\)
where
\begin{equation}
B_j=\begin{bmatrix}
A_{1j}   &A_{2j}   & \cdots & A_{qj}      \\
I & \cdots &0 &0\\
\vdots & \ddots & \vdots \\
0      & \cdots & I & 0
\end{bmatrix} 
\end{equation}

\(I\) is the identity matrix with the same dimension as \(A_{ij}\). We represent different model\(f_j\) with different matrix\(B_j\) and \(B_jZ_{t-1}=f_j(Z_{t-1})\). Then this flattened VAR(q) model is one special case of the regression model with input to be \(Z_{t-1}\) and output to be \(Z_t\). The input is distributed dependently, then the consistency analysis is guaranteed by theorem \ref{thm:dependent}. In this section with use the VAR(4) model and the coefficient matrices for different models are generated by the algorithm in paper \cite{VARparam}. 

Table \ref{tab:exp} showed the performance of our algorithms on these baseline simulations. The neural network for all the experiments we run in the training step of the algorithm is a fully-connected network with 2 hidden layers with 256 units and ReLU activation function. The change points \(\tau_j\) are simulated by a random sample function. The ``distance between change points" means the range that random function used to generate the distance between the change points and N means the number of change points. For each setting of distance range and change point number N, we generate the data set for 10 repetitions and run the two-step algorithm. Then we demonstrate the performance of the algorithm by \(\bar{d}_{\tau-\hat{\tau}}=\frac{1}{10}\sum_{i=1}^{10} d_i(\tau-\hat{\tau}) \) where each \(d_i(\tau-\hat{\tau})=\frac{1}{N}\sum_{j=1}^{N}|\tau_j-\hat{\tau}_j|\) (Here \(\tau_j,\hat{\tau}_j\) is the location of jth true change point and estimated jth change point and the we will choose the closest estimated change point to \(\tau_j\) to be the estimated jth change point) for ith repetition, \(\bar{d}_{|N-\hat{N}|}=\frac{1}{10}\sum_{i=1}^{10} |N-\hat{N}_i|\) where each \(\hat{N}_i\) is the number of estimated change points for ith repetition and Prop(\(\hat{N}=N\)) is the proportion of the 10 repetitions that we have \(\hat{N}_i=N\).

\begin{table*}[t]
  \caption{Several Basic experiments}
  \label{tab:exp}
  \centering
    \renewcommand{\arraystretch}{1.2} % Adjust row spacing for better readability
\begin{tabular}{@{}lccccc@{}} 
    \toprule
    Model & Distance Between CPs & N & \(\bar{d}_{\tau-\hat{\tau}}\) & \(\bar{d}_{|N-\hat{N}|}\) & Proportion (\(\hat{N}=N\)) \\
    \midrule
    Feed-Forward Network Data & 300--1500 & 30 & 17.68 & 0.0 & 1.0 \\

    Random Forest & 300--1500 & 30 & 17.30 & 0.0 & 1.0 \\
    VAR(4) Flatten & 900--1500 & 30 & 20.89 & 0.0 & 1.0 \\
    \bottomrule
\end{tabular}
\end{table*}
In this section, we will also run simulations to illustrate the consistency of the locations and number of change points detected for different types of datasets. The first comparison is the independent vs dependent data input. The input for the independent setting is generated by i.i.d normal distribution N(0,I) and dependent input is generated by VAR(1) model and normalize the two types of input to the same scale for each \(X_i\), the model function \(f_j\) is the same feed-forward network structure as displayed in the previous simulation for both model setting. To easily compare the performance of the algorithm under the two types of data input, we implement a narrow distance(200-300 time points) for change points.

\begin{table*}[t]
\caption{Independent vs dependent input}
\label{tab:indvsdep}
\centering
\renewcommand{\arraystretch}{1.2}
\begin{tabular}{@{}lccccc@{}} 
  \toprule
  model& Distance between cps & N & \(\bar{d}_{\tau-\hat{\tau}}\) & \(\bar{d}_{|N-\hat{N}|}\) & Prop(\(\hat{N}=N\)) \\ 
  \midrule
    Independent input data &  200-300 & 30&18.76 & 0.3 & 0.7\\ 
  Dependent input data  &  200-300 & 30&88.14 &7.7 &0.0\\ 
  \bottomrule
\end{tabular}
\end{table*}
We can observe from the results table \ref{tab:indvsdep} that the algorithm will get better performance(lower distance and higher proportion of correct detection) on the independent data input than dependent input which is illustrated by the theorem \ref{thm:independent} and \ref{thm:dependent}.

In theorems \ref{thm:independent} and \ref{thm:subgaussian}, we also demonstrate the difference of algorithm performance given different model output distribution. The table below is the comparison between Nonsubgaussian data vs subgaussian output. The model function for both data sets is generated by feed-forward network structure. Elements in \(W_i\) for both data sets are still from N(0,1) and elements \(W_{ij}\) are sparse(0.1)*N(0,1) for subgaussian and sparse(0.1)*Uniform(0,1) for nonsubgaussian data output. To get a justifiable result, we fix the signal for both data sets. For both two types of data models, we set the model change signal to be constant:\(E_{p(X)}[||f_i(X)-f_{i-1}(X)||^2]=50, i=1,2,...,N\).

We can observe from this table \ref{tab:subvsnonsub} that the algorithm will get better performance given the subgaussian data output.
\begin{table*}[t]
\caption{Nonsubgaussian vs subgaussian}
\label{tab:subvsnonsub}
\centering
\renewcommand{\arraystretch}{1.2}
\begin{tabular}{@{}lccccc@{}} 
  \toprule
  model& Distance between cps & N & \(\bar{d}_{\tau-\hat{\tau}}\) & \(\bar{d}_{|N-\hat{N}|}\) & Prop(\(\hat{N}=N\)) \\ 
  \midrule
    Non-subgaussian data & 200-300 & 60&29.76 & 3.5 & 0.2\\ 
  Subgaussian data &  200-300 &60& 15.95 &0.2 &0.8\\ 
  \bottomrule
\end{tabular}
\end{table*}

In some cases, we aim to estimate the change points with greater precision by minimizing the distance between the true change points and the estimated ones. For datasets with finite \(l\)-moment of the output for sufficiently large \(l\), decreasing the test window size \(T_0\) can improve the precision of change point localization. This behavior is demonstrated in Table \ref{tab:precise} .The input, output, and model function remain the same as those in the feed-forward network setting described in the first simulation. From the table \ref{tab:precise}, it is significant that reducing the window size \(T_0\) leads to more accurate estimation of change point locations.
\begin{table}[t]
\caption{Precise train}
\label{tab:precise}
\centering
\renewcommand{\arraystretch}{1.2}
\begin{tabular}{@{}lcccc@{}}
  \toprule
  model& Distance between cps & N & \(\bar{d}_{\tau-\hat{\tau}}\) & \(T_0\)\\ 
  \midrule
    First train &  300-1500 & 10& 20.46 & 50\\ 
  Precise train  &  300-1500 &10 & 8.43  &20\\ 
  \bottomrule
\end{tabular}
\end{table}

Finally, in the theorem, the assumption that the signal should be significantly larger than the standard deviation of the error term \(\sigma\) is indispensable, as demonstrated in the proof provided in the supplementary section. The Figure \ref{fig:sigma} illustrates the relationship between the magnitude of \(\sigma\) and the algorithm's performance. This is evaluated through three metrics: the average distance between true and estimated change points \(\bar{d}_{\tau-\hat{\tau}}\),  the average difference between the number of detected and true change points \(\bar{d}_{|N-\hat{N}|}\)  and the proportion of instances where the exact number of change points is detected Prop(\(\hat{N}=N\)).
\begin{figure*}[!t]
    \centering
    \subfloat[\(\bar{d}_{\tau-\hat{\tau}}\) vs \(\sigma\)]{\includegraphics[width=0.34\textwidth]{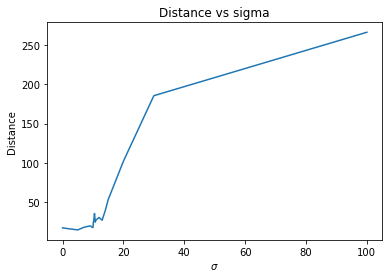}} 
    \subfloat[\(\bar{d}_{|N-\hat{N}|}\) vs \(\sigma\)]{\includegraphics[width=0.34\textwidth]{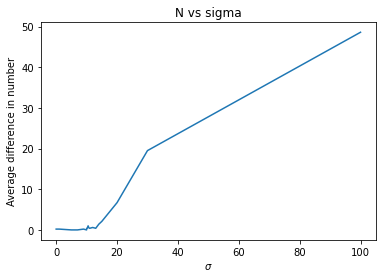}} 
    \subfloat[Prop(\(\hat{N}=N\)) vs \(\sigma\)]{\includegraphics[width=0.34\textwidth]{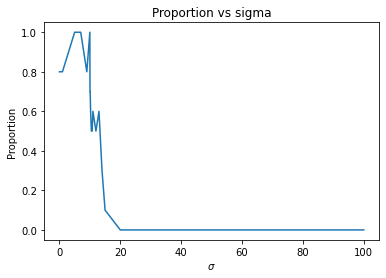}} 
    \caption{Performance metrics vs \(\sigma\)}
    \label{fig:sigma}
\end{figure*}
We observe that as the magnitude of \(\sigma\) increases, the performance of the algorithm deteriorates significantly. This highlights the importance of the assumptions regarding the signal's magnitude in ensuring the algorithm's effectiveness.
\subsection{Real Data}
We also present the results of applying our detection strategy to a variety of real-world datasets.

\begin{figure*}[!t]
    \centering
    \subfloat[The genes mutations of Cancer patients]{\includegraphics[width=0.34\textwidth]{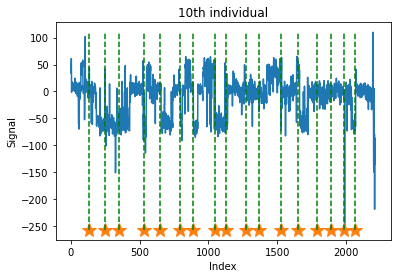}} 
    \subfloat[log difference of GDP]{\includegraphics[width=0.34\textwidth]{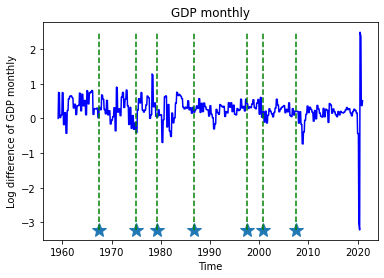}} 
    \subfloat[DJIA Stock market index]{\includegraphics[width=0.34\textwidth]{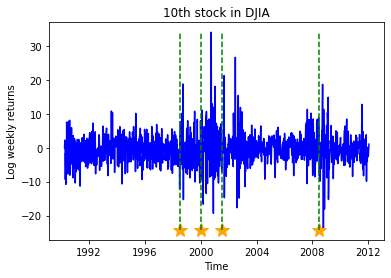}} 
    \caption{Real data}
    \label{fig:real}
\end{figure*}

The first dataset consists of data from 43 individuals with bladder tumors, where each value represents the copy number of specific gene mutations for each segment of each individual. The dataset contains 2216 segments \cite{Bleakley2011TheGF}. Using the same detection strategy as in the VAR model, we apply a window size of \(T_0=50\simeq \sqrt{2216}\). Our two-step algorithm detected 17 change points in this dataset. Figure (a) in Figure \ref{fig:real} illustrates the number of mutation copies related to bladder cancer, with the detected change points marked by stars and vertical dashed lines.

The second dataset contains the log weekly returns for 29 stocks comprising the Dow Jones Industrial Average (DJIA). The dataset spans 1138 time points, from April 1990 to January 2012, where each time point represents one week \cite{JSSv062i07}.For this dataset, we use a window size of \(T_0=30\simeq \sqrt{1138}\). Our two-step algorithm successfully detected changes near the years 1998, 2000, 2002, and 2008, corresponding to significant events such as the 1997 financial crisis, the collapse of the technology bubble, the stock market downturn of 2002, and the financial crisis of 2008. Figure (c) in Figure \ref{fig:real} shows the log weekly returns for the 10th stock used to calculate the DJIA index, with detected change points highlighted.

The third dataset, sourced from \cite{GDP}, consists of 743 time points, where each time point represents a month. The 1-dimensional output represents the log difference in monthly GDP, while the 122-dimensional input comprises various economic indicators, including stock market data, real personal income, and US treasury bond returns. This dataset spans several key recessions in U.S. economic history. Our algorithm detected change points corresponding to events such as the 1973 oil crisis, the 1980 recession, the dot-com bubble of 2000, and the great recession of 2008, among others.

\subsection{Comparison with commonly used algorithms}
We have also compared our algorithm with popular algorithms including KLCPD in paper \cite{klcpd}, Sparsified Binary Segmentation in paper\cite{sparsified} and double CUSUM in paper\cite{doublecusum}. 

The data we used are generated from three different synthetical data models VAR(5), Non-linear VAR, and Lotka-Volterra model with multiple species.

The VAR(5) model is defined same as in section \ref{sec:exp} equation \eqref{eq:VAR} with dimension 200.

Non-linear VAR model introduced in paper \cite{NonlinearVAR} is defined as:
\begin{equation}
x_t=Ax_{t-1}+\sum_{i=0}^{q}\Lambda_i\rho_i(f_{rt-i} +\epsilon_t)
\end{equation}
where 
\begin{equation}
f_t=\Gamma_1f_{t-1}+\Gamma_2f_{t-2}+\epsilon_t^f
\end{equation}
Here \(\Gamma_j,\epsilon_t^f\) is the coefficient matrix and error term for VAR(2) model \(f_t\) and A and \(\epsilon_t\) is the coefficient matrix and error term for Non-linear VAR model \(x_t\). \(\Lambda_i\) is the coefficient matrix for non-linear lag terms in Non-linear VAR model and \(\rho_i\) is the corresponding non-linear function. In experiment, we set q=6, dimension of \(f_t\) to be 10, dimension of \(x_t\) to be 200. We also have the frequency parameter \(r=3\) here to skip direct connection between the consecutive components \(f_t,f_{t-1}\).

Lotka-Volterra model introduced in paper \cite{grangercausality} (displayed as LV in the table \ref{tab:comparison}) with multiple species are generated by the following differential equations:
\begin{subequations}
\label{LV}
\begin{align}
\frac{dx^i}{dt}=\alpha x^i-\beta x^i \sum_{j\in Pa(x^i)} y^j-\alpha(x^i)^2, 1\leq i\leq p \label{LVA} \\
\frac{dy^j}{dt}=\delta y^j\sum_{k\in Pa(y^j)} x^k-\gamma(y^j),1\leq j\leq p \label{LVB}
\end{align}
\end{subequations}
where \(x^i\) corresponds to the population sizes of prey species and \(y^j\) is the population sizes of predator species, \(\alpha,\beta,\delta, \gamma\) are the fixed parameters controlling strengths of interactions, and the \(Pa(x^i),Pa(y^j)\) are the sets of Granger-causes of \(x^i,y^j\).

Here we define 3 types of distance. The regular distance \(d_{\tau-\hat{\tau}}\) we used in section \ref{sec:exp}, the sum version Hausdorff distance 
\begin{equation}
d\_hau\_w=\max_{\tau_i\in T} \sum_{\substack{\hat{\tau}_i \in \hat{T}|\tau_i-\hat{\tau}_j|<=\min_k|\tau_k-\hat{\tau}_j| } }|\tau_i-\hat{\tau}_j|
\end{equation}
the product version Hausdorff distance 
\begin{equation}
d\_hau\_prod=\max_{\tau_i\in T} \prod_{\substack{\hat{\tau}_i \in \hat{T}  |\tau_i-\hat{\tau}_j|<=\min_k|\tau_k-\hat{\tau}_j| } }|\tau_i-\hat{\tau}_j|
\end{equation}
where \(T,\hat{T}\) are the set for true change points and detected change points. For the comparison, the precision and recall will also be presented. We use these two modified versions of Hausdorff distances because many algorithms can get decent results in distance and accuracy when detecting change points. But they usually suggest multiple estimation around the true change point and it is not applicable in real world when we don't know which estimated change point should be selected finally. By introducing these new metrics, we can penalize such behavior and make sure one algorithm will not propose too many estimations around the true change points when it gets good performance in these metrics and we can get a better understanding of the performance and robustness of different change point detection algorithms.

In the experiments, the distances between change points are randomly generated from the uniform distribution ranging from 1200 to 1500.

The results of the comparison experiments are displayed in the table \ref{table:comparison}. In the table, the five metrics are displayed for every combination of detection methods and datasets. Our methods beat the 3 competitive methods in two of the datasets except for Nonlinear VAR data where Double Cusum has better performance under some metrics. 

\begin{table*}[!t] % Use table* to span both columns
\centering
\caption{Comparison of Methods on Various Models}
\label{tab:comparison}
\begin{center}
  \renewcommand{\arraystretch}{1.2}
\setlength{\tabcolsep}{3pt}
\begin{tabular}{@{}lcccccc@{}}
  \toprule
  Method& Data & $\bar{d}_{\tau-\hat{\tau}}$ & d\_hau\_w & d\_hau\_prod & precision & recall\\ 
  \midrule
  ML model & Nonlinear VAR &118.16&\textbf{134.0} & 31.7&0.675 & 0.70\\
    Sparsified Binary Segmentation & Nonlinear VAR & 157.94 &  400.2& 284.5 & 0.362& \textbf{0.975}\\ 
  Double Cusum & Nonlinear VAR & \textbf{90.00} &130.20 &\textbf{8.9}&\textbf{0.704}&0.925\\
  KLCPD & Nonlinear VAR & 499.36 & 302.7& 1108.5 &0.09& 0.05\\
  \midrule
  ML model & LV& \textbf{87.84} &  \textbf{132.90} & \textbf{48.2} & \textbf{0.58} & \textbf{0.75}\\
 Sparsified Binary Segmentation & LV & 480.84 & 391.0 &5803.7& 0.11 & 0.65\\
  Double Cusum & LV & 561.25 & 850.0& 273.70 &0.0& 1.0\\
  KLCPD & LV & 1601.09 & 209.4& inf &0.0& 0.0\\
  \midrule
    ML model & VAR& \textbf{20.05} &  \textbf{32.5}& \textbf{32.5} & \textbf{0.90} & \textbf{0.875}\\
 Sparsified Binary Segmentation & VAR & 480 & 391.0 &5803.7& 0.11 & 0.65\\
  Double Cusum & VAR & 561.25 & 850.0& 273.70 &0.0& 1.0\\
  KLCPD & VAR & 862.64 & 284.0& inf &0.09& 0.1\\
  \bottomrule

\end{tabular}
\end{center}
\label{table:comparison}
\end{table*}

\section{Conclusion and Discussion}
In this paper, we introduced a two-step algorithmic framework for offline change point detection using feed-forward neural networks. This approach effectively combines a training step for estimating model parameters and a detection step for identifying change points based on an error criterion. We established theoretical guarantees for the number and locations of change points detected by our method, ensuring its reliability under specific assumptions. These guarantees include precise bounds on the distances between estimated and true change points, as well as the conditions under which the detected change points accurately match the true ones.

Our extensive experiments validated the performance of the algorithm across various simulated and real-world datasets. For instance, in simulated datasets based on feed-forward networks, random forests, and vector autoregressive models, our algorithm demonstrated robust detection capabilities under diverse settings. Moreover, experiments with real-world data—including genetic mutation copy number datasets, stock market returns for the Dow Jones Industrial Average (DJIA), and GDP-related economic indicators—further highlighted the applicability and accuracy of the proposed approach. These results showcased the algorithm's ability to detect meaningful change points in practical scenarios, such as identifying shifts during economic recessions or significant market events.

In this paper, we also compare our algorithm with methods including KLCPD and Sparsified Binary Segmentation on different multivariate scenarios.

While the detection strategy is generalizable to other neural network architectures, such as recurrent neural networks (RNNs), convolutional neural networks (CNNs), and long short-term memory networks (LSTMs), the current theoretical framework does not extend to these architectures. Future work could address this limitation by establishing theoretical guarantees for these network types, potentially broadening the scope of applications.

Another promising direction for future research involves refining the method for dependent data cases. Specifically, by incorporating assumptions such as martingale properties or Markov chain dependencies, the algorithm could achieve more precise change point detection with reduced window sizes. This adjustment would enhance its applicability to time series data with complex dependencies, such as financial or biological systems.

Additionally, exploring the scalability of the algorithm to larger datasets or online scenarios represents another avenue for advancement. Optimizing computational efficiency and adapting the algorithm for real-time detection could make it suitable for applications requiring immediate responses, such as anomaly detection in streaming data or high-frequency financial data analysis.

In summary, the proposed two-step framework not only offers a theoretically grounded and experimentally validated solution for offline change point detection but also paves the way for future research to expand its versatility and robustness in broader contexts.

\section{Annotation and Assumption}
In this supplement section, we will provide proof of the theorems in the consistency analysis section.

Before heading into the proof of the main theorems, as we said in theoretical analysis section\ref{sec:thm}, results in \cite{Schmidt_Hieber_2020,ma2022theoretical} for independent and dependent data are leveraged regarding how well the estimated feed-forward network approximates the true underlying \(f_j\). The first section of the supplement will be focused on the notations, assumptions, and the lemmas we need for the main proof. The second section will be mainly on the proof of theorem \ref{thm:single} and the third section will be proof of theorem \ref{thm:independent}, \ref{thm:subgaussian}, and \ref{thm:dependent}.

\section{Proof}
\subsection{Proof of Theorem\ref{thm:single}}
\label{sec:proof1}
In this section, the proof of subgaussian and dependent cases will be covered in the next section for multiple change point cases. Also, the proof of existence for the peak in E(t) will be covered in the next section. Moreover, the lemma above keeps the output of the neural network to be one dimension. In our algorithm, the output,  however, is the h dimension. To mitigate this difference, we consider each feed-forward network we use as \(\hat{f}=(\hat{f}_1,...,\hat{f}_h)\) where each \(\hat{f}_k \in \mathcal{F}(L,\boldsymbol{p},s,F)\) represents a network defined the same as in the lemma and similarly \(f_0=(f_0^1,...,f_0^h)\) where\(f_0^k \in \mathcal{G}(q,\boldsymbol{d},\boldsymbol{t},\boldsymbol{\beta},K)\).  Then 
\begin{equation}
\mathbb{E}_{f_0}[||\hat{f}(X)-f_0(X)||^2]=\sum_{k=1}^h \mathbb{E}_{f_0}[(\hat{f}_k(X)-f_0^k(X))^2]
\end{equation}
where each term in the sum is controlled by the lemma \ref{lem01} and the sum will also be controlled with multiplication by h.

For the single change point case in theorem\ref{thm:single}, \(\tau\) is the estimated change point where \(\tau=arg\max_t(Error(t))\). Here we first assume \(T_0\gg C|\tau-\tau^*|\) for some constant C. 

If \(\tau^*>\tau\):
\begin{equation}
E(\tau)-E(\tau^*)=
\sum_{i=\tau}^{\tau+T_0}||\hat{f}^1_{\tau}(X_i)-Y_i||^2-
\sum_{i=\tau^*}^{\tau^*+T_0}||\hat{f}^2_{\tau^*}(X_i)-Y_i||^2\geq 0
\end{equation}

Where \(\hat{f}^1_{\tau} \in arg\min_{f\in \mathcal{F}(L,p,s,F)}\sum_{i\leq \tau}(Y_i-f(X_i))^2\) and \(\hat{f}^2_{\tau^*} \in arg\min_{f\in \mathcal{F}(L,p,s,F)}\sum_{i\leq \tau^*}(Y_i-f(X_i))^2\)\\
And then
\begin{multline}
\sum_{i=\tau}^{\tau^*}||\hat{f}^1_{\tau}(X_i)-Y_i||^2+\sum_{i=\tau^*}^{\tau+T_0}||\hat{f}^1_{\tau}(X_i)-Y_i||^2
-\sum_{i=\tau^*}^{\tau+T_0}||\hat{f}^2_{\tau^*}(X_i)-Y_i||^2-\sum_{i=\tau+T_0}^{\tau^*+T_0}||\hat{f}^2_{\tau^*}(X_i)-Y_i||^2\geq 0\\
\sum_{i=\tau}^{\tau^*}||\hat{f}^1_{\tau}(X_i)-Y_i||^2+\sum_{i=\tau^*}^{\tau+T_0}(||\hat{f}^1_{\tau}(X_i)-Y_i||^2-||\hat{f}^2_{\tau^*}(X_i)-Y_i||^2)
-\sum_{i=\tau+T_0}^{\tau^*+T_0}||\hat{f}^2_{\tau^*}(X_i)-Y_i||^2\geq 0
\end{multline}
Here let \(I_1=\sum_{i=\tau}^{\tau^*}||\hat{f}^1_{\tau}(X_i)-Y_i||^2\), \(I_2=\sum_{i=\tau^*}^{\tau+T_0}(||\hat{f}^1_{\tau}(X_i)-Y_i||^2-||\hat{f}^2_{\tau^*}(X_i)-Y_i||^2)\) and \(I_3=\sum_{i=\tau+T_0}^{\tau^*+T_0}||\hat{f}^2_{\tau^*}(X_i)-Y_i||^2\)\\
For \(I_1\):
\begin{multline}
I_1 \leq \sum_{i=\tau}^{\tau^*}(||\hat{f}^1_{\tau}(X_i)-f_0(X_i)||^2+||Y_i-f_0(X_i)||^2)\\
\leq (\tau^*-\tau)\mathbb{E}_{f_0}[(\hat{f}^1_{\tau}(X)-f_0(X)^2] +o(\frac{1}{\tau^*-\tau}) +\sum_{\tau}^{\tau^*}||Y_i-f_0(X_i)||^2 \\
\overset{Theorem\ref{lem01}}{\leq} C_1 \phi_T L \log^2T(\tau^*-\tau)+h\sigma^2(\tau^*-\tau)
\end{multline}
For \(I_2\):
\begin{multline}
I_2 =\sum_{i=\tau^*}^{\tau+T_0}(||\hat{f}^1_{\tau}(X_i)-Y_i||^2-||\hat{f}^2_{\tau^*}(X_i)-Y_i||^2)
\overset{\text{Cauchy Inequity}}{\leq} \\
\sum_{i=\tau^*}^{\tau+T_0}(||\hat{f}^1_{\tau}(X_i)-\hat{f}^2_{\tau^*}(X_i)||)(||\hat{f}^1_{\tau}(X_i)+\hat{f}^2_{\tau^*}(X_i)-2Y_i||) \\
\overset{\text{Cauchy Inequity}}{\leq} \sqrt{T_0E_{f_1}[||\hat{f}^1_{\tau}(X)-\hat{f}^2_{\tau^*}(X)||^2]+o(\frac{1}{T_0})}\\
\sqrt{\bigg(\sum_{i=\tau^*}^{\tau+T_0}||\hat{f}^1_{\tau}(X_i)-f_0(X_i)||^2+||\hat{f}^2_{\tau^*}(X_i)-f_0(X_i)||^2
+2||Y_i-f_0(X_i)||^2}\bigg) \\
 \overset{Theorem\ref{lem01}}{\leq} \sqrt{C_2 T_0\phi_T L \log^2T} \\
\sqrt{\bigg(\sum_{i=\tau^*}^{\tau+T_0}||\hat{f}^1_{\tau}(X_i)-f_0(X_i)||^2+||\hat{f}^2_{\tau^*}(X_i)-f_0(X_i)||^2 
 +2||Y_i-f_0(X_i)||^2}\bigg) \\
\overset{Minkovsky Inequity}{\leq} \sqrt{C_2 T_0\phi_T L \log^2T} \\ (\sqrt{\sum_{i=\tau^*}^{\tau+T_0}||\hat{f}^1_{\tau}(X_i)-f_0(X_i)||^2+||\hat{f}^2_{\tau^*}(X_i)-f_0(X_i)||^2} \\
+\sqrt{\sum_{i=\tau^*}^{\tau+T_0}2||Y_i-f_1(X_i)||^2+2||f_0(X_i)-f_1(X_i)||^2}) \\
\leq \sqrt{C_2 T_0\phi_T L \log^2T} (\sqrt{2T_0\mathbb{E}_{f_1}[||\hat{f}^2_{\tau^*}(X)-f_0(X)||^2]+o(\frac{1}{T_0})} \\
+\sqrt{\sum_{i=\tau^*}^{\tau+T_0}2||Y_i-f_1(X_i)||^2+2||f_0(X_i)-f_1(X_i)||^2}) \\
\overset{Thoerem\ref{lem01}}{\leq} \sqrt{C_2 T_0\phi_T L \log^2T} \bigg(\sqrt{2C_2T_0\phi_T L \log^2T} 
+\sqrt{\sum_{i=\tau^*}^{\tau+T_0}2T_0h\sigma^2+2M_2T_0+o(\frac{1}{T_0})}\bigg) \\
\leq 2C_2T_0\phi_T L \log^2T+\sqrt{2C_2\phi_TL\log^2Th}T_0\sigma \\
+\sqrt{2C_2\phi_TL\log^2TM_2}T_0 \leq C_4(M_2,C_2,L)T_0 \phi_T \log^2 T
\end{multline}
Where \(C_4\) is a constant given \(M_2,C_2\) and L.

For \(I_3\):
\begin{multline}
I_3=\sum_{i=\tau+T_0}^{\tau^*+T_0}||\hat{f}^2_{\tau^*}(X_i)-Y_i||^2 \geq \sum_{i=\tau+T_0}^{\tau^*+T_0}(||f_0(X_i)-f_1(X_i)||^2\\
-||\hat{f}^2_{\tau^*}(X_i)-f_0(X_i)||^2-||f_1(X_i)-Y_i||^2)\\
 \geq (\tau^*-\tau)\mathbb{E}_{f_0}[(f_0(X)-f_1(X))^2]-(\tau^*-\tau)\mathbb{E}_{f_1}[(\hat{f}^2_{\tau^*}(X)-f_1(X))^2]\\
 -(\tau^*-\tau)\mathbb{E}[\epsilon^2]+o(\frac{1}{\tau^*-\tau})\\
 \overset{Theorem\ref{lem01}}{\geq} (\tau^*-\tau)(M_1-h\sigma^2-C_3\phi_T L\log^2T)   
\end{multline}

By the previous equation \(I_1+I_2 \geq I_3\)
\begin{multline}
 C_1 \phi_T L \log^2T(\tau^*-\tau)+h\sigma^2(\tau^*-\tau) 
 +C_4(M_2,C_2,L)T_0^2 \phi_T \log^2T \\
 \geq (\tau^*-\tau)(M_1-h\sigma^2-C_3\phi_T L\log^2T) \\
 |\tau^*-\tau| \leq \frac{C_4(M_2,C_2,L)T_0 \phi_T \log^2T}{M_1-2h\sigma^2-(C_1+C_3)\phi_T L\log^2T}
\end{multline}
Similarly, if \(\tau>\tau^*\):
\begin{equation}
Error(\tau)-Error(\tau^*)=
\sum_{i=\tau}^{\tau+T_0}||\hat{f}^1_{\tau}(X_i)-Y_i||^2-\sum_{i=\tau^*}^{\tau^*+T_0}||\hat{f}^2_{\tau^*}(X_i)-Y_i||^2\geq 0
\end{equation}
Where \(\hat{f}^1_{\tau} \in arg\min_{f\in \mathcal{F}(L,p,s,F)}\sum_{\tau^*\leq i\leq \tau}(Y_i-f(X_i))^2\) and \(\hat{f}^2_{\tau^*} \in arg\min_{f\in \mathcal{F}(L,p,s,F)}\sum_{i\leq \tau^*}(Y_i-f(X_i))^2\)\\
And then
\begin{multline}
\sum_{i=\tau^*+T_0}^{\tau+T_0}||\hat{f}^1_{\tau}(X_i)-Y_i||^2+\sum_{i=\tau}^{\tau^*+T_0}||\hat{f}^1_{\tau}(X_i)-Y_i||^2\\
-\sum_{i=\tau}^{\tau^*+T_0}||\hat{f}^2_{\tau^*}(X_i)-Y_i||^2-\sum_{i=\tau^*}^{\tau}||\hat{f}^2_{\tau^*}(X_i)-Y_i||^2\geq 0\\
-\sum_{i=\tau^*}^{\tau}||\hat{f}^2_{\tau^*}(X_i)-Y_i||^2+ 
\sum_{i=\tau}^{\tau^*+T_0}(||\hat{f}^1_{\tau}(X_i)-Y_i||^2-||\hat{f}^2_{\tau^*}(X_i)-Y_i||^2)\\
+\sum_{i=\tau^*+T_0}^{\tau+T_0}||\hat{f}^1_{\tau}(X_i)-Y_i||^2\geq 0
\end{multline}
Here let \(I_1=\sum_{i=\tau^*+T_0}^{\tau+T_0}||\hat{f}^1_{\tau}(X_i)-Y_i||^2\), \(I_2=\sum_{i=\tau}^{\tau^*+T_0}(||\hat{f}^1_{\tau}(X_i)-Y_i||^2-||\hat{f}^2_{\tau^*}(X_i)-Y_i||^2)\) and \(I_3=\sum_{i=\tau^*}^{\tau}||\hat{f}^2_{\tau^*}(X_i)-Y_i||^2\) 

For \(I_1\):
\begin{multline}
I_1 \leq \sum_{i=\tau^*+T_0}^{\tau+T_0}(||\hat{f}^1_{\tau}(X_i)-f_1(X_i)||^2+||Y_i-f_1(X_i)||^2)\\
\leq (\tau^*-\tau)\mathbb{E}_{f_1}[(\hat{f}^1_{\tau}(X)-f_1(X)^2] 
+o(\frac{1}{\tau-\tau^*}) +\sum_{\tau^*+T_0}^{\tau+T_0}||Y_i-f_1(X_i)||^2 \\
\overset{Theorem\ref{lem01}}{\leq} C_1 \phi_{\tau-\tau^*} L \log^2(\tau-\tau^*)(\tau-\tau^*)+h\sigma^2(\tau-\tau^*)
\leq C_1\min_{i=1,...,L}p_i\log^2(\tau-\tau^*)+h\sigma^2(\tau-\tau^*) \\
\leq h\sigma^2(\tau-\tau^*)+(\tau-\tau^*)o(1)
\end{multline}
For \(I_2\):
\begin{multline}
I_2=\sum_{i=\tau}^{\tau^*+T_0}(||\hat{f}^1_{\tau}(X_i)-Y_i||^2-||\hat{f}^2_{\tau^*}(X_i)-Y_i||^2)\\
\leq \sum_{i=\tau}^{\tau^*+T_0}(||\hat{f}^1_{\tau}(X_i)-f_1(X_i)||^2+||f_1(X_i)-Y_i||^2)-\sum_{i=\tau}^{\tau^*+T_0}||\hat{f}^2_{\tau^*}(X_i)-Y_i||^2 \\
\leq T_0\mathbb{E}_{f_1}[(\hat{f}^1_{\tau}(X)-f_1(X))^2]+T_0h\sigma^2+o(\frac{1}{T_0})\\
-\sum_{i=\tau}^{\tau^*+T_0}||(f_0(X_i)-f_1(X_i))+(\hat{f}^2_{\tau^*}(X_i)-f_0(X_i))+(f_1(X_i)-Y_i)||^2 \\
\leq T_0\mathbb{E}_{f_1}[(\hat{f}^1_{\tau}(X)-f_1(X))^2]+T_0h\sigma^2\\
-\sum_{i=\tau}^{\tau^*+T_0}(||f_0(X_i)-f_1(X_i)||^2-||\hat{f}^2_{\tau^*}(X_i)-f_0(X_i)||^2-||f_1(X_i)-Y_i||^2) \\
\overset{Theorem\ref{lem01}}{\leq} T_0C_1 \phi_{\tau-\tau^*} L \log^2(\tau-\tau^*)+2T_0h\sigma^2+ \\
C_1 \phi_{T} L \log^2(T)-T_0M_2\\
\leq T_0\min_{i=1,...,L}p_i\frac{log^2(\tau-\tau^*)}{\tau-\tau^*}+2T_0h\sigma^2-T_0M_2+C_1 \phi_{T} L \log^2(T)
\end{multline}
For \(I_3\):
\begin{multline}
I_3=\sum_{i=\tau^*}^{\tau}||\hat{f}^2_{\tau^*}(X_i)-Y_i||^2\\
=\sum_{i=\tau^*}^{\tau}||(\hat{f}^2_{\tau^*}(X_i)-f_0(X_i))+(f_0(X_i)-f_1(X_i))+(f_1(X_i)-Y_i)||^2\\
\geq \sum_{i=\tau^*}^{\tau}(||f_0(X_i)-f_1(X_i)||^2-||\hat{f}^2_{\tau^*}(X_i)-f_0(X_i)||^2-||f_1(X_i)-Y_i||^2)\\
 \geq (\tau-\tau^*)\mathbb{E}_{f_0}[(f_0(X)-f_1(X))^2]\\
 -(\tau-\tau^*)\mathbb{E}_{f_1}[(\hat{f}^2_{\tau^*}(X)-f_1(X))^2]-(\tau-\tau^*)\mathbb{E}[\epsilon^2]+o(1)\\
 \overset{Theorem\ref{lem01}}{\geq} (\tau-\tau^*)(M_1-h\sigma^2-C_3\phi_T L\log^2T)
\end{multline}
By the previous equation \(-I_3+I_2+I_1\geq 0\)
\begin{multline}
(\tau-\tau^*)o(1)+h\sigma^2(\tau-\tau^*)+T_0\min_{i=1,...,L}p_i\frac{\log^2(\tau-\tau^*)}{\tau-\tau^*}\\
+2T_0h\sigma^2+C_1 \phi_{T} L \log^2(T)-T_0M_2 \\
\geq (\tau-\tau^*)(M_1-h\sigma^2-C_3\phi_T L\log^2T) \\
|\tau-\tau^*| \leq \\
\frac{2T_0h\sigma^2+C_1 \phi_{T} L \log^2(T)-T_0M_2+T_0\min_{i=1,...,L}p_i\frac{\log^2(\tau-\tau^*)}{\tau-\tau^*}}{M_1-2h\sigma^2-C_3\phi_T L\log^2T-o(1)}
\end{multline}
Here, if \(|\tau-\tau^*|\geq \sqrt{T_0}\), there \(\exists C_5\) constant, \(T_0\min_{i=1,...,L}p_i\frac{\log^2(\tau-\tau^*)}{\tau-\tau^*}\leq C_5 T_0\)
\begin{multline}
|\tau-\tau^*| \leq \frac{2T_0h\sigma^2-T_0M_2+C_5T_0+C_1 \phi_{T} L \log^2(T)}{M_1-2h\sigma^2-C_3\phi_T L\log^2T} \\
\leq \frac{C_5T_0+C_1 \phi_{T} L \log^2(T)}{M_1-2h\sigma^2-C_3\phi_T L\log^2T}
\end{multline}
Here if \(|\tau-\tau^*|>CT_0\), if \(\tau^*>\tau\).
\begin{multline}
E(\tau)-E(\tau^*)=
\sum_{i=\tau}^{\tau+T_0}||\hat{f}^1_{\tau}(X_i)-Y_i||^2-\sum_{i=\tau^*}^{\tau^*+T_0}||\hat{f}^2_{\tau^*}(X_i)-Y_i||^2 \\
\overset{Theorem\ref{lem01}}{\leq} 2T_0C\phi_TL\log^2T+2T_0h\sigma^2-M_1T_0 < 0
\end{multline}
If \(\tau>\tau^*\)
\begin{multline}
E(\tau)-E(\tau^*)=
\sum_{i=\tau}^{\tau+T_0}||\hat{f}^1_{\tau}(X_i)-Y_i||^2-\sum_{i=\tau^*}^{\tau^*+T_0}||\hat{f}^2_{\tau^*}(X_i)-Y_i||^2 \\
\overset{Theorem\ref{lem01}}{\leq} 2T_0C\phi_{T_0}L\log^2(T_0)+2T_0h\sigma^2-M_1T_0 < 0
\end{multline}
A contradiction of \(Error(\tau)\geq Error(\tau^*)\).\\
In all, there exists constant \(C_5, C\), with probability \(1-\frac{C}{T_0}\) 
\begin{equation}
|\tau^*-\tau|\leq \frac{C_5T_0}{M_1-2h\sigma^2}
\end{equation}
According to the proof in section \ref{sec:proof1}, the probability 
\begin{equation}
P\bigg(|\sum_{i=\tau}^{\tau+T_0}||\hat{f}^1_{\tau}(X_i)-f_1(X_i)||^2 \\
-T_0\mathbb{E}[||\hat{f}^1_{\tau}(X_i)-f_1(X_i)||^2]|>\delta\bigg)\leq \frac{C(\delta)}{T_0}
\end{equation}
given \(\forall \delta\), C is a constant depending on \(\delta\). For other approximations in the proof, the technique is similar. This probability will be specified clearly in the proof of multiple change point cases in the section \ref{subsec:proof2} given the different assumptions of the model and data.

\subsection{Proof of theorems \ref{thm:independent}, \ref{thm:subgaussian} and \ref{thm:dependent}}
\label{subsec:proof2}

Let \(T_0\) be the test window size. \(\hat{\tau}_i\) denotes the \(i^{th}\) estimate change point. By the proof of a single change point, 
\begin{multline}
 |\hat{\tau}_i-\tau_i|
 \leq \max( \frac{C_{f_i}(M2_i)T_0 \phi_{|\tau_{i+1}-\tau_i|}\log^2|\tau_{i+1}-\tau_i|}{M1_i-2h\sigma^2-C\phi_{|\tau_{i+1}-\tau_i|} L\log^2|\tau_{i+1}-\tau_i|}, \\
 \frac{C_5T_0+C_1 \phi_{|\tau_{i+1}-\tau_i|} L \log^2(|\tau_{i+1}-\tau_i|)}{M1_i-2h\sigma^2-C_3\phi_{|\tau_{i+1}-\tau_i|} L\log^2|\tau_{i+1}-\tau_i|},\sqrt{T_0})\\
\leq \max \bigg( \frac{C_{f_i}(M2_i)T_0 \phi_{T}\log^2T}{M1_i-2h\sigma^2}, \frac{C_5T_0+C_1 \phi_{T} L \log^2(T)}{M1_i-2h\sigma^2} ,\sqrt{T_0}\bigg)\leq C\frac{C_5}{M1_i-2h\sigma^2}T_0
\end{multline}
By the lemma \ref{lem3}, with probability \(1-\frac{C}{T_0^{p-1}}\) if X is independent of each other and \(1-\frac{C}{T_0}\)if X is dependent of each other, the above inequity will hold.

If X variable is subgaussian and independent of each other, by the lemma \ref{lem1}, the probability will be \(1-Ce^{-C_1T_0}\). 

Inside the algorithm \ref{alg:detect}, define \(T_0=\sqrt[\rho]{T}=o(T)\), for each peak found, define the peak window as \(C_6T_0=C\frac{C_5}{M1_*-2h\sigma^2}T_0\) and \(\pi=\frac{M1_*}{2}-2h\sigma^2\) then inside the interval \([\min(t-C_6T_0,0),(t+C_6T_0)]\) when detecting peak at time t, there is one and only one change point with high probability depending on the type of input and model because windows of the nearby two peaks detected \([\min(t'-C_6T_0,0),(t'+C_6T_0)]\) which containing the relative true change points will not intersect with it. Also, the next detection window will be at least one detection window size \(C_6T_0\) away from the previous window, and will not detect the previous change point for the second time.

Then we need also to prove near each change point, the algorithm will detect a peak in the error function.
Given any change point \(\tau_i,1\leq i\leq N\). Let\(\hat{f}_{\tau_i-2T_0}\in arg\min_{f\in \mathcal{F}(L,p,s,F)}\sum_{k=\tau_{i-1}}^{\tau_i-2T_0}||Y_k-f(X_k)||^2\), \(\hat{f}_{\tau_i-T_0}\in arg\min_{f\in \mathcal{F}(L,p,s,F)}\sum_{k=\tau_{i-1}}^{\tau_i-T_0}||Y_k-f(X_k)||^2\), \(\hat{f}_{\tau_i}\in arg\min_{f\in \mathcal{F}(L,p,s,F)}\sum_{k=\tau_{i-1}}^{\tau_i}||Y_k-f(X_k)||^2\), \(\hat{f}_{\tau_i+T_0}\in arg\min_{f\in \mathcal{F}(L,p,s,F)}\sum_{k=\tau_i}^{\tau_i+T_0}||Y_k-f(X_k)||^2\) and \(\hat{f}_{\tau_i+2T_0}\in arg\min_{f\in \mathcal{F}(L,p,s,F)}\sum_{k=\tau_i}^{\tau_i+2T_0}||Y_k-f(X_k)||^2\). To detect a peak near the true change point, the equation below holds.
\begin{multline}
  \min\left(Error(\tau_i)-Error(\tau_i-T_0),Error(\tau_i)-Error(\tau_i+T_0)\right) \\
  \gg \max\bigg(|Error(\tau_i-T_0)-Error(\tau_i-2T_0)| \\
  ,|Error(\tau_i+T_0)-Error(\tau_i+2T_0)|\bigg)
\end{multline}
Which means 

\begin{multline}
  \min(\sum_{k=\tau_i}^{\tau_i+T_0}||Y_k-\hat{f}_{\tau_i}(X_k)||^2-\sum_{k=\tau_i-T_0}^{\tau_i}||Y_k-\hat{f}_{\tau_i-T_0}(X_k)||^2,\\
  \sum_{k=\tau_i}^{\tau_i+T_0}||Y_k-\hat{f}_{\tau_i}(X_k)||^2-\sum_{k=\tau_i+T_0}^{\tau_i+2T_0}||Y_k-\hat{f}_{\tau_i+T_0}(X_k)||^2)\\
  \gg \max(|\sum_{k=\tau_i-T_0}^{\tau_i}||Y_k-\hat{f}_{\tau_i-T_0}(X_k)||^2 
  -\sum_{k=\tau_i-2T_0}^{\tau_i-T_0}||Y_k-\hat{f}_{\tau_i-2T_0}(X_k)||^2 |\\
  ,|\sum_{k=\tau_i+T_0}^{\tau_i+2T_0}||Y_k-\hat{f}_{\tau_i+T_0}(X_k)||^2 
  -\sum_{k=\tau_i+2T_0}^{\tau_i+3T_0}||Y_k-\hat{f}_{\tau_i+2T_0}(X_k)||^2|)
\end{multline}
Let \(J_1=\sum_{k=\tau_i}^{\tau_i+T_0}||Y_k-\hat{f}_{\tau_i}(X_k)||^2-\sum_{k=\tau_i-T_0}^{\tau_i}||Y_k-\hat{f}_{\tau_i-T_0}(X_k)||^2\), \(J_2=\sum_{k=\tau_i}^{\tau_i+T_0}||Y_k-\hat{f}_{\tau_i}(X_k)||^2-\sum_{k=\tau_i+T_0}^{\tau_i+2T_0}||Y_k-\hat{f}_{\tau_i+T_0}(X_k)||^2\), \(J_3=|\sum_{k=\tau_i-T_0}^{\tau_i}||Y_k-\hat{f}_{\tau_i-T_0}(X_k)||^2-\sum_{k=\tau_i-2T_0}^{\tau_i-T_0}||Y_k-\hat{f}_{\tau_i-2T_0}(X_k)||^2 |\) and \(J_4=|\sum_{k=\tau_i+T_0}^{\tau_i+2T_0}||Y_k-\hat{f}_{\tau_i+T_0}(X_k)||^2-\sum_{k=\tau_i+2T_0}^{\tau_i+3T_0}||Y_k-\hat{f}_{\tau_i+2T_0}(X_k)||^2|\).

For \(J_1\):
\begin{multline}
 J_1=\sum_{k=\tau_i}^{\tau_i+T_0}||Y_k-\hat{f}_{\tau_i}(X_k)||^2-\sum_{k=\tau_i-T_0}^{\tau_i}||Y_k-\hat{f}_{\tau_i-T_0}(X_k)||^2\\
 \geq \sum_{k=\tau_i}^{\tau_i+T_0}(||f_{i+1}(X_k)-f_i(X_k)||^2-||Y_k-f_{i+1}(X_k)||^2)
 -\sum_{k=\tau_i}^{\tau_i+T_0}||f_i(X_k)-\hat{f}_{\tau_i}(X_k)||^2 \\
 -\sum_{k=\tau_i-T_0}^{\tau_i}(||Y_k-f_i(X_k)||^2+||f_i(X_k)-\hat{f}_{\tau_i-T_0}||^2)\\
 \overset{Theorem\ref{lem01}}{\geq} o(1)+M1_iT_0-C_1T_0\phi_{\tau_i-\tau_{i-1}}L\log^2(\tau_i-\tau_{i-1})\\
 -C_2T_0\phi_{\tau_i-\tau_{i-1}-T_0}L\log^2(\tau_i-\tau_{i-1}-T_0)-2T_0h\sigma^2
\end{multline}

For \(J_2\):
\begin{multline}
J_2=\sum_{k=\tau_i}^{\tau_i+T_0}||Y_k-\hat{f}_{\tau_i}(X_k)||^2-\sum_{k=\tau_i+T_0}^{\tau_i+2T_0}||Y_k-\hat{f}_{\tau_i+T_0}(X_k)||^2\\
\geq \sum_{k=\tau_i}^{\tau_i+T_0}(||f_{i+1}(X_k)-f_i(X_k)||^2-||Y_k-f_{i+1}(X_k)||^2) 
-\sum_{k=\tau_i}^{\tau_i+T_0}||f_i(X_k)-\hat{f}_{\tau_i}(X_k)||^2\\
-\sum_{k=\tau_i+T_0}^{\tau_i+2T_0}(||Y_k-f_{i+1}(X_k)||^2+||f_{i+1}(X_k)-\hat{f}_{\tau_i+T_0}(X_k)||^2)\\
\overset{Theorem\ref{lem01}}{\geq} o(1)+M1_iT_0-C_1T_0\phi_{\tau_i-\tau_{i-1}}L\log^2(\tau_i-\tau_{i-1})
-C_2T_0\phi_{T_0}L\log^2(T_0)-2T_0h\sigma^2
\end{multline}

For \(J_3\):
\begin{multline}
 J_3=|\sum_{k=\tau_i-T_0}^{\tau_i}||Y_k-\hat{f}_{\tau_i-T_0}(X_k)||^2 
 -\sum_{k=\tau_i-2T_0}^{\tau_i-T_0}||Y_k-\hat{f}_{\tau_i-2T_0}(X_k)||^2|\\
 \leq \sum_{k=\tau_i-T_0}^{\tau_i}(||Y_k-f_i(X_k)||^2+||f_i(X_k)-\hat{f}_{\tau_i-T_0}(X_k)||^2) \\
 +\sum_{k=\tau_i-2T_0}^{\tau_i-T_0}(||Y_k-f_i(X_k)||^2+||f_i(X_k)-\hat{f}_{\tau_i-2T_0}(X_k)||^2)\\
 \overset{Theorem\ref{lem01}}{\leq} o(1)+C_1T_0\phi_{\tau_i-\tau_{i-1}-2T_0}L\log^2(\tau_i-\tau_{i-1}-2T_0)\\
 +C_2T_0\phi_{\tau_i-\tau_{i-1}-T_0}L\log^2(\tau_i-\tau_{i-1}-T_0)+2T_0h\sigma^2
\end{multline}
For \(J_4\):
\begin{multline} J_4=|\sum_{k=\tau_i+T_0}^{\tau_i+2T_0}||Y_k-\hat{f}_{\tau_i+T_0}(X_k)||^2-\sum_{k=\tau_i+2T_0}^{\tau_i+3T_0}||Y_k-\hat{f}_{\tau_i+2T_0}(X_k)||^2|\\
 \leq \sum_{k=\tau_i+T_0}^{\tau_i+2T_0}(||Y_k-f_{i+1}(X_k)||^2+||f_{i+1}(X_k)-\hat{f}_{\tau_i+T_0}(X_k)||^2) \\
 +\sum_{k=\tau_i+2T_0}^{\tau_i+3T_0}(||Y_k-f_{i+1}(X_k)||^2+||f_{i+1}(X_k)-\hat{f}_{\tau_i+2T_0}(X_k)||^2)\\
 \overset{Theorem\ref{lem01}}{\leq} o(1)+2T_0h\sigma^2+C_1T_0\phi_{T_0}L\log^2T_0+C_2T_0\phi_{2T_0}L\log^2(2T_0)
\end{multline}
By the previous equation, \(T_0\ll T \leq \tau_i-\tau_{i-1}\), \(\min(J_1,J_2)=J_2,\max(J_3,J_4)=J_4\). To let \(\min(J_1,J_2)\gg \max(J_3,J_4)\) which is \(J_2\gg J_4\).
\begin{multline}
o(1)+M1_iT_0-2T_0h\sigma^2-C_1T_0\phi_{\tau_i-\tau_{i-1}}L\log^2(\tau_i-\tau_{i-1})
-C_2T_0\phi_{T_0}L\log^2(T_0)\\
\gg o(1)+2T_0h\sigma^2+C_1T_0\phi_{T_0}L\log^2T_0+C_2T_0\phi_{2T_0}L\log^2(2T_0)
\end{multline}
To satisfy the condition above, 
\begin{equation}
\begin{split}
T_0\gg \frac{o(1)}{M1_i-4h\sigma^2-C\phi_{T_0}L\log^2(T_0)}
\end{split}
\end{equation}
This will be satisfied as \(T_0\to \infty\) when \(T_{sum}\to \infty\) and then the algorithm will detect a peak near the true change point with probability \(1-\frac{C}{T_0}\) (or \(1-\frac{C}{T_0^{p-1}}\) and \(1-e^{-CT_0}\)) for some constant C depending on the type of input and model. And when \(\pi=\frac{M1_*}{2}-2h\sigma^2\), \(J_2-J_4>\pi\) thus finishing this part of proof.

Then, we start to compute the probability of the approximation. By assumption and then we can finish the proof for these separate cases. 

For independent variables, for each changepoint, with probability at least \(1-\frac{C}{T_0^{p-1}}\), the algorithm will detect the changepoint. Then, with probability at least \((1-\frac{C}{T_0^{p-1}})^N\), the algorithm will detect all N changepoints and exactly these N changepoints and the distance between the estimated change point and the true one will be controlled.
\begin{multline}
(1-\frac{C}{T_0^{p-1}})^N\geq 1-\frac{CN}{T_0^{p-1}} \geq 1-\frac{CT_{sum}}{TT_0^{p-1}}\geq 1-\frac{C_0T_{sum}}{T^{1+\rho(p-1)}} \\
\geq 1-C_0T_{sum}^{-\frac{1}{\rho(p-1)}} \underset{T_{sum}\to \infty}{\to} 1
\end{multline}

For independent and subgaussian variables, for each changepoint, with probability at least \(1-Ce^{-C_1T_0}\), the algorithm will detect the changepoint. Then, with probability at least \((1-Ce^{-C_1T_0})^N\), the algorithm will detect all N changepoints and exactly these N changepoints and the distance between the estimated change point and the true one will be controlled.
\begin{multline}
(1-Ce^{-C_1T_0})^N\geq 1-CNe^{-C_1T_0} \geq 1-C_0\frac{T_{sum}}{Te^{C_1T_0}} \\
\geq 1-\frac{CT_{sum}}{C_2\log(T_{sum})T_{sum}}=1-\frac{C_4}{\sqrt{\log(T_{sum})}} \to 1
\end{multline}

For dependent variables, for each changepoint, with probability at least \(1-\frac{C}{T_0}\), the algorithm will detect the changepoint. Then, with probability at least \((1-\frac{C}{T_0})^N\), the algorithm will detect all N changepoints and exactly these N changepoints and the distance between the estimated change point and the true one will be controlled.

\begin{multline}
(1-\frac{C}{T_0})^N\geq 1-\frac{CN}{T_0} \geq 1-\frac{CT_{sum}}{TT_0}\geq 1-\frac{C_0T_{sum}}{T^{1+\rho}} \\
\geq 1-\frac{C_0T_{sum}}{T_{sum}^{\frac{\rho+1}{1+g\rho}}} \geq 1-C_0T_{sum}^{-\frac{(1-g)\rho}{g\rho++1}} \underset{T_{sum}\to \infty}{\to} 1
\end{multline}
Then we finish the proof.

\section{reference}

\bibliographystyle{IEEEtran} % Or any other style like plain, alpha, etc.
\bibliography{reference}

\section{Supplement}
\subsection{\(\beta\)-H\"older smoothness}
\label{subsec:betaholder}
First, we need some assumptions and also demonstrate the features of the model functions \(f_j\) which are composed of several functions: \(f_j=g_q\circ g_{q-1}\circ ...\circ g_1\circ g_0\). We need to define a type of function called \(\beta\)-H\"older smoothness whos partial derivatives exist up to \(\lfloor \beta \rfloor\) order and are bounded and the partial derivatives of order \(\lfloor \beta\rfloor\) are \(\beta-\lfloor \beta \rfloor\) H\"older. Then define the ball of \(\beta\)-H\"older smoothness functions with radius K as:

\begin{multline}
     \mathcal{C}_r^{\beta}(D,K)=\{f:D\subset R^r \to R; \sum_{\boldsymbol{\alpha}:|\boldsymbol{\alpha}|<\beta}||\partial^{\boldsymbol{\alpha}}f||_{\infty}+\\
     \sum_{\boldsymbol{\alpha}:|\boldsymbol{\alpha}|=\lfloor\beta\rfloor}\underset{x,y\in D;x\neq y}{\sup}\frac{|\partial^{\boldsymbol{\alpha}}f(x)-\partial^{\boldsymbol{\alpha}}f(y)|}{|x-y|_{\infty}^{\beta-\lfloor \beta\rfloor}} \leq K \}  
\end{multline}

where \(\partial^{\boldsymbol{\alpha}}=\partial^{\alpha_1}...\partial^{\alpha_r}\) with \(\boldsymbol{\alpha}=(\alpha_1,...,\alpha_r) \in \mathbb{N}^r\) where \(\partial^{\alpha_i}\) denotes the \(\alpha_i\) partial derivative on the \(i^{th}\) dimension and \(|\boldsymbol{\alpha}|:=|\boldsymbol{\alpha}|_1\). We assume \(g_{ij}\) is \(\beta\)-H\"older smoothness then we define a function space to be:

\begin{multline}
  \mathcal{G}(q,\boldsymbol{d},\boldsymbol{t},\boldsymbol{\beta},K):=\{ f=g_q\circ g_{q-1}\circ ...\circ g_1\circ g_0: \\
  g_i=(g_{ij})_j:[a_i,b_i]^{d_i}\to [a_{i+1},b_{i+1}]^{d_{i+1}},\\
   g_{ij} \in  \mathcal{C}_{t_i}^{\beta_i}([a_i,b_i]^{t_i},K) \text{ where } |a_i|\leq K,|b_i|\leq K \}  
\end{multline}
where \(\boldsymbol{d}:=(d_0,...,d_{q+1}),\boldsymbol{t}:=(t_0,...,t_{q}),\boldsymbol{\beta}:=(\beta_0,...,\beta_{q})\). This function space is the domain of our model function.

\subsection{Neural network Function space}
\label{subsec:neuralfunction}
Then we start to define the function space of network functions. First, we define a network function space with bounded parameters as:
\begin{multline}
  \mathcal{F}(L,\boldsymbol{p}):=\{ f:\mathbb{R}^{p_0}\to \mathbb{R}^{p_{L+1}}, \\
  x \to f(x)=W_L\sigma_{v_L}W_{L-1}\sigma_{v_{L-1}}...W_1\sigma_{v_1}W_0x: \\ \underset{j=0,...,L}{\max}||W_j||_{\infty}\vee |v_j|_{\infty}\leq 1 \} 
\end{multline}
Where \(\boldsymbol{p}=(p_0,...,p_{L+1})\in \mathbb{N}^{L+2}\), L is the number of layers of the network, \(W_j\) is the \(p_{j+1} \times p_j\) matrix, \(v_j \in R^{p_j}\) and \(\sigma_{v_j}(x)=\max(0,x-v_j)\) is the ReLU function. Then \(f: \mathbb{R}^{p_0}\to \mathbb{R}^{p_{L+1}}\) defined above is a neural network with ReLU activation. Here \(p_0=l\) and \(p_{L+1}=h\).

Let \(|||f|_{\infty}||_{\infty}\) be sup-norm of the function \(x\to |f(x)|_{\infty}\). Add the sparse condition to the parameters and define the network function space as:
\begin{multline}
  \mathcal{F}(L,\boldsymbol{p},s):=\mathcal{F}(L,\boldsymbol{p},s,F):= \\
  \{ f\in \mathcal{F}(L,\boldsymbol{p}):\sum_{j=0}^L||W_j||_{0}+|v_j|_{0}\leq s, |||f|_{\infty}||_{\infty} \leq F \}  
\end{multline}
Then start to approximate the convergence rate. First, let
\begin{equation}
 \beta_i^* := 	\prod_{i+1}^q (\beta_l \wedge 1)  
\end{equation}
And define convergence rate as:
\begin{equation}
\phi_n := \underset{i=0,...,q}{\max}n^{-\frac{2\beta_i^*}{\beta_i^*+t_i}}    
\end{equation}
where n is the size of the training set where the network is trained.

\subsection{Lemma}
The next two lemmas in paper \cite{Schmidt_Hieber_2020,ma2022theoretical} will be the key to our proof for the main theorems. The model function space and the network function space will be defined as above.
\begin{lemma}
\label{lem01}
Let \(f_0 \in \mathcal{G}(q,\boldsymbol{d},\boldsymbol{t},\boldsymbol{\beta},K)\) and define a network function space \(\mathcal{F}(L,\boldsymbol{p},s,F)\).Let \(Y_i=f_0(X_i)+\epsilon_i\) as defined by model\ref{model} where X is independent and\(\hat{f} \in arg\min_{f\in \mathcal{F}(L,\boldsymbol{p},s,F)}\sum_{i=1}^n(Y_i-f(X_i))^2 \)be an
empirical risk minimizer and satisfied conditions below:
\\
(i)\(F\geq \max(K,1)\)
\\
(ii)\(\sum_{i=0}^q\log_2(4t_i\vee 4\beta_i)\log_2n\leq L \lesssim n\phi_n\)
\\
(iii)\(n\phi_n \lesssim \min_{i=1,...,L}p_i\)
\\
(iv)\(s\simeq n\phi_n\log n\)
\\
There exists a constant C' only depending on q,\(\boldsymbol{d},\boldsymbol{t},\boldsymbol{\beta}\), F such that
\[ R(\hat{f}-f_0)=\mathbb{E}_{f_0}[(\hat{f}(X)-f_0(X))^2] \leq C'\phi_{n}L\log^2n\]
\end{lemma}

\begin{lemma}

Let \(f_0 \in \mathcal{G}(q,\boldsymbol{d},\boldsymbol{t},\boldsymbol{\beta},K)\) and define a network function space \(\mathcal{F}(L,\boldsymbol{p},s,F)\).Let \(Y_i=f_0(X_i)+\epsilon_i\) as defined by equation\ref{eq:problem} in section \ref{sec:problem} where X is dependent and\(\hat{f} \in arg\min_{f\in \mathcal{F}(L,\boldsymbol{p},s,F)}\sum_{i=1}^n(Y_i-f(X_i))^2 \)be an
empirical risk minimizer and satisfied conditions below:
\\
(i)\(F\geq \max(K,1)\)
\\
(ii)\(\sum_{i=0}^q\log_2(4t_i\vee 4\beta_i)\log_2n\leq L \lesssim n\phi_n\)
\\
(iii)\(n\phi_n \lesssim \min_{i=1,...,L}p_i\)
\\
(iv)\(s\simeq n\phi_n\log n\)
\\
There exists a constant C' only depending on q,\(\boldsymbol{d},\boldsymbol{t},\boldsymbol{\beta}\), F such that
\[ R(\hat{f}-f_0)=\mathbb{E}_{f_0}[(\hat{f}(X)-f_0(X))^2] \leq C'\phi_{n}L\log^6n\]
\end{lemma}

For the proof of our theorems, we also need 3 common concentration inequalities:
\begin{lemma}
\label{lem1}
Given \(X_1, ...,X_n \) i.i.d subgaussian, then for \(\forall \delta>0\), there exists constant \(C_1,C_2\) greater than 0.
\[
P\left( |\bar{X}-\mu|>\delta  \right)\leq C_1e^{-C_2n}
\]
where \(\mu=\mathbb{E}[X_i]\) and \(\bar{X}=\frac{1}{n}\sum_{i=1}^nX_i\)
\end{lemma}

\begin{lemma}
\label{lem2}
Given \(X_1, ...,X_n \) independent and from the same distribution, then if \(\mathbb{E}(|X_i|^p)<\infty \), there exists a constant C>0,
\[
P\left( |\bar{X}-\mu|>\delta  \right)\leq \frac{C}{n^{p-1}} 
\]
where \(\mu=\mathbb{E}[X_i]\) and \(\bar{X}=\frac{1}{n}\sum_{i=1}^nX_i\)
\end{lemma}

\begin{lemma}
\label{lem3}
Given \(X_1, ...,X_n \) dependent and from the same distribution, then if \(|cov(X_i,X_j)|<c<1 \) and \(\mathbb{E}(|X_i|^2)<\infty \), there exists a constant C>0,
\[
P\left( |\bar{X}-\mu|>\delta  \right)\leq \frac{C}{n} 
\]
where \(\mu=\mathbb{E}[X_i]\) and \(\bar{X}=\frac{1}{n}\sum_{i=1}^nX_i\)
\end{lemma}

\subsection{Additional figures}
\begin{figure*}[!t]
    \centering
    \subfloat[qqplot for nonsubgaussian data]{\includegraphics[width=0.4\textwidth]{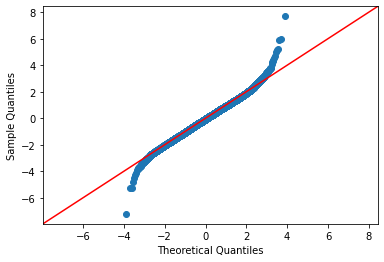}} 
    \subfloat[qqplot for subgaussian data]{\includegraphics[width=0.4\textwidth]{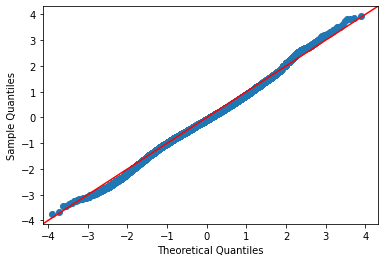}} 
    \caption{QQplot}
    \label{fig:qq}
\end{figure*}

The figure \ref{fig:qq} demonstrates the normality of the data output. The left qqplot is for nonsubgaussian data and right is for subgaussian data.

\vfill

\end{document}